\documentclass{article}

\usepackage{pdfpages}
\usepackage{microtype}
\usepackage{graphicx}
\usepackage{subfigure}
\usepackage{booktabs}

\usepackage{hyperref}

\usepackage[accepted]{icml2018}

\icmltitlerunning{Comparison-Based Random Forests}

\usepackage{multirow}
\usepackage{graphicx}
\usepackage{setspace}
\usepackage{amsmath, amsthm, amssymb}

\usepackage{dsfont}
\usepackage{algorithm}
\usepackage{algorithmic}
\usepackage{float}

\usepackage{mathtools}

\usepackage{enumitem}

\renewcommand{\algorithmicrequire}{\textbf{Input:}}
\renewcommand{\algorithmicensure}{\textbf{Output:}}

\newcommand{\Xc}{\mathcal{X}}

\DeclareMathOperator{\Diam}{diam}
\DeclareMathOperator{\Expec}{\mathbb{E}}
\DeclareMathOperator{\Indic}{\mathds{1}}
\DeclareMathOperator{\Proba}{\mathbb{P}}

\newcommand{\bigo}[1]{O\left(#1\right)}
\newcommand{\ceil}[1]{\left\lceil#1\right\rceil}
\newcommand{\condproba}[2]{\Proba\left (#1\middle|#2\right )}
\newcommand{\diam}[1]{\Diam\left(#1\right)}
\newcommand{\Defeq}{\vcentcolon =}
\newcommand{\Density}{f}

\newcommand{\Dist}{\delta}
\newcommand{\Dim}{d}
\newcommand{\dist}[2]{\Dist\left(#1,#2\right)}
\newcommand{\expec}[1]{\Expec\left[#1\right]}
\newcommand{\floor}[1]{\left\lfloor#1\right\rfloor}
\newcommand{\indic}[1]{\Indic_{#1}}
\newcommand{\Inputspace}{\Xc}
\newcommand{\littleo}[1]{\mathrm{o}\left(#1\right)}

\newcommand{\norm}[1]{\left\lVert#1\right\rVert}
\newcommand{\normalcondset}[2]{\bigl\{#1\mid#2\bigr\}}
\newcommand{\proba}[1]{\Proba\left (#1\right )}
\newcommand{\Reals}{\mathbb{R}}

\newcommand{\ra}[1]{\renewcommand{\arraystretch}{#1}}

\theoremstyle{definition}
\newtheorem{assumption}{Assumption}[section]
\newtheorem{definition}{Definition}
\theoremstyle{plain}
\newtheorem{theorem}{Theorem}[section]
\newtheorem{proposition}{Proposition}[section]

\begin{document}

\twocolumn[
\icmltitle{Comparison-Based Random Forests}

\begin{icmlauthorlist}
\icmlauthor{Siavash Haghiri}{TUE}
\icmlauthor{Damien Garreau}{MPG}
\icmlauthor{Ulrike von Luxburg}{TUE,MPG}
\end{icmlauthorlist}

\icmlaffiliation{TUE}{Department of Computer Science, University of T{\"u}bingen, Germany}
\icmlaffiliation{MPG}{Max Planck Institute for Intelligent Systems, T{\"u}bingen, Germany}

\icmlcorrespondingauthor{Siavash Haghiri}{haghiri@informatik.uni-tuebingen.de}

\icmlkeywords{Random forests, Ordinal information, Triplet comparisons}

\vskip 0.3in
]
\printAffiliationsAndNotice{}
\begin{abstract}
  Assume we are given a set of items from a general metric space, but we neither have access to the representation of the data nor to the distances between data points.
  Instead, suppose that we can actively choose a triplet of items $(A,B,C)$ and ask an oracle whether item~$A$ is closer to item~$B$ or to item~$C$. 
  In this paper, we propose a novel random forest algorithm for regression and classification that relies only on such triplet comparisons. In the theory part of this paper, we 
  establish sufficient conditions for the consistency of  such a forest. In a set of comprehensive experiments, we then demonstrate that the proposed random forest is efficient both for classification and regression. In particular, it is even competitive with other methods that have direct access to the metric representation of the data. 
\end{abstract}

\section{Introduction}

Assume we are given a set of items from a general metric space $(\Xc, \delta)$, but we neither have access to the representation of the data nor to the distances between data points. 
Instead, we have access to an oracle that we can actively ask a \textbf{triplet comparison}: 
given any triplet of items $(x_i,x_j,x_k)$ in the metric space $\Xc$, is it true that 
\[
\delta(x_i,x_j) < \delta(x_i,x_k) \;\;\;\text{?}
\]
Such a comparison-based framework has become popular in recent years, for example in the context of crowd-sourcing applications \citep{tamuz2011adaptively,heikinheimo2013crowd,ukkonen2015crowdsourced}, and more generally, whenever humans are supposed to give feedback or when constructing an explicit distance or similarity function is difficult \citep{wilber2015learning,zhang2015jointly,wah2015learning,balcan2016Learning,kleindessner2017kernel}. 

In the present work, we consider classification and regression problems in a comparison-based setting where we are given the labels $y_1,\ldots, y_n$ of unknown objects $x_1,\ldots, x_n$, and we can actively query triplet comparisons between objects.
An indirect way to solve such problems is to first construct an ``ordinal embedding'' of the data points in a (typically low-dimensional) Euclidean space that satisfies the set of triplet comparisons, and then to apply standard machine learning methods to the Euclidean data representation. 
However, this approach is often not satisfactory because this new representation necessarily introduces distortions in the data.
Furthermore, all existing ordinal embedding methods are painfully slow, even on moderate-sized datasets \citep{agarwal2007generalized,van2012stochastic,terada2014local}. 
In addition, one has to estimate the embedding dimension, which is a challenging task by itself~\citep{kleindessner2015dimensionality}.

As an alternative, we solve the classification/regression problems by a new random forest algorithm that requires only triplet comparisons. Standard random forests \citep{biau2016random} are one of the most popular and successful classification/regression algorithms in Euclidean spaces \citep{fernandez2014we}. However, they \emph{heavily} rely on the underlying vector space structure. In our comparison-based setting we need a completely different tree building strategy. We use the recently described comparison tree \citep{haghiri2017comparison} for this purpose
(which in Euclidean cases would be distantly related to linear decision trees \citep{kane2017active,ezra2017nearly,kane2017near}). 
A comparison-based random forest (CompRF) consists of a collection of comparison trees built on the training set.

We study the proposed CompRF both from a theoretical and a practical point of view. 
In Section~\ref{sec:analysis}, we give sufficient conditions under which a slightly simplified variant of the comparison-based forest is statistically consistent. 
In Section~\ref{sec:experiments}, we apply the CompRF to various datasets. In the first set of experiments we compare our random forests to traditional CART forests on Euclidean data. In the second set of experiments, the distances between objects are known while their representation is missing. 
Finally, we consider a case in which only triplet comparisons are available. 

\section{Comparison-Based Random Forests}

\begin{figure*}[ht]
    \vskip 0.2in
    \centering
    \hfill
    \begin{subfigure}
    \centering
    \includegraphics[width=.19\linewidth]{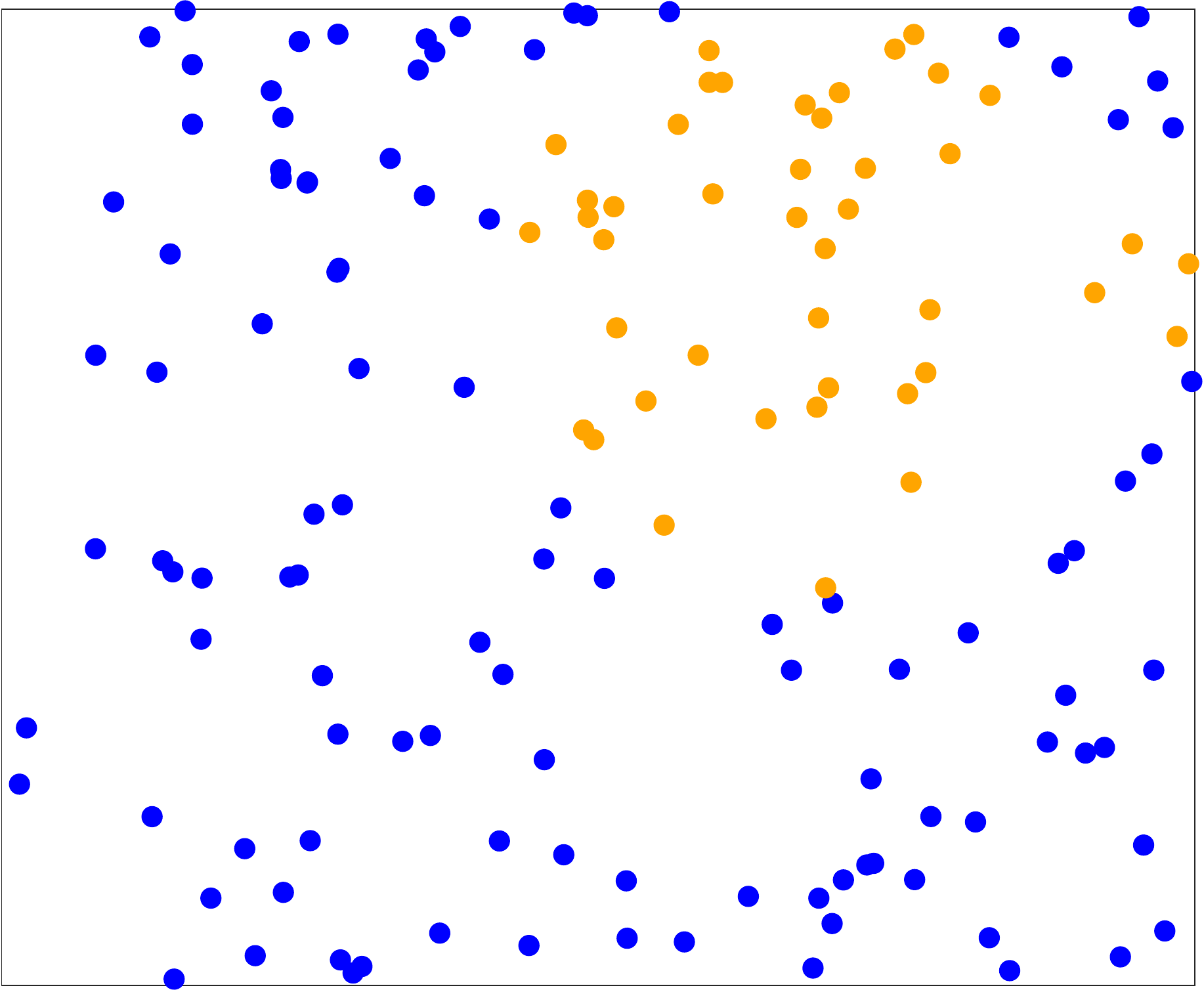}
    \end{subfigure}\hfill
    \begin{subfigure}
    \centering
    \includegraphics[width=.19\linewidth]{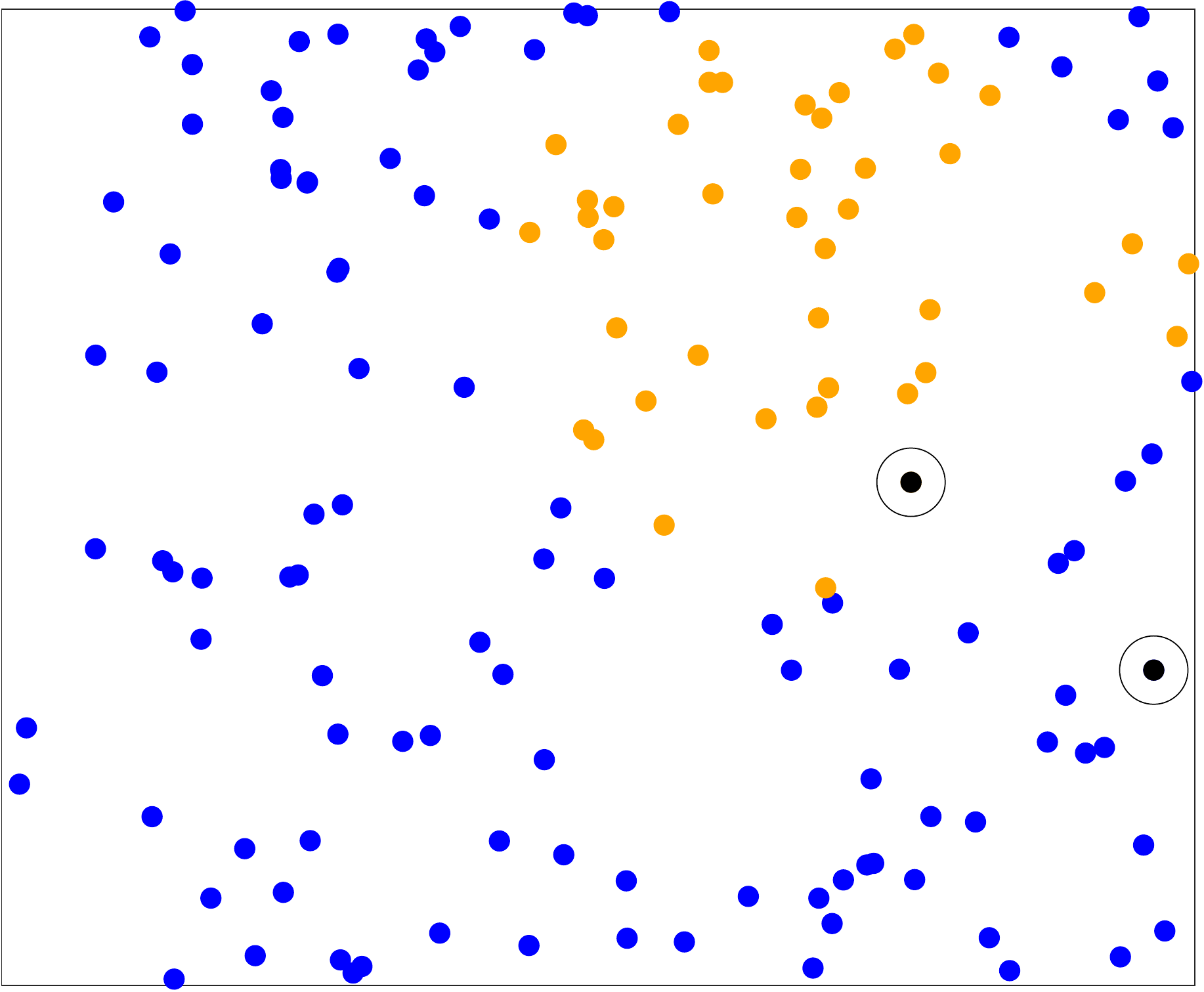}
    \end{subfigure}\hfill
    \begin{subfigure}
    \centering
    \includegraphics[width=.185\linewidth]{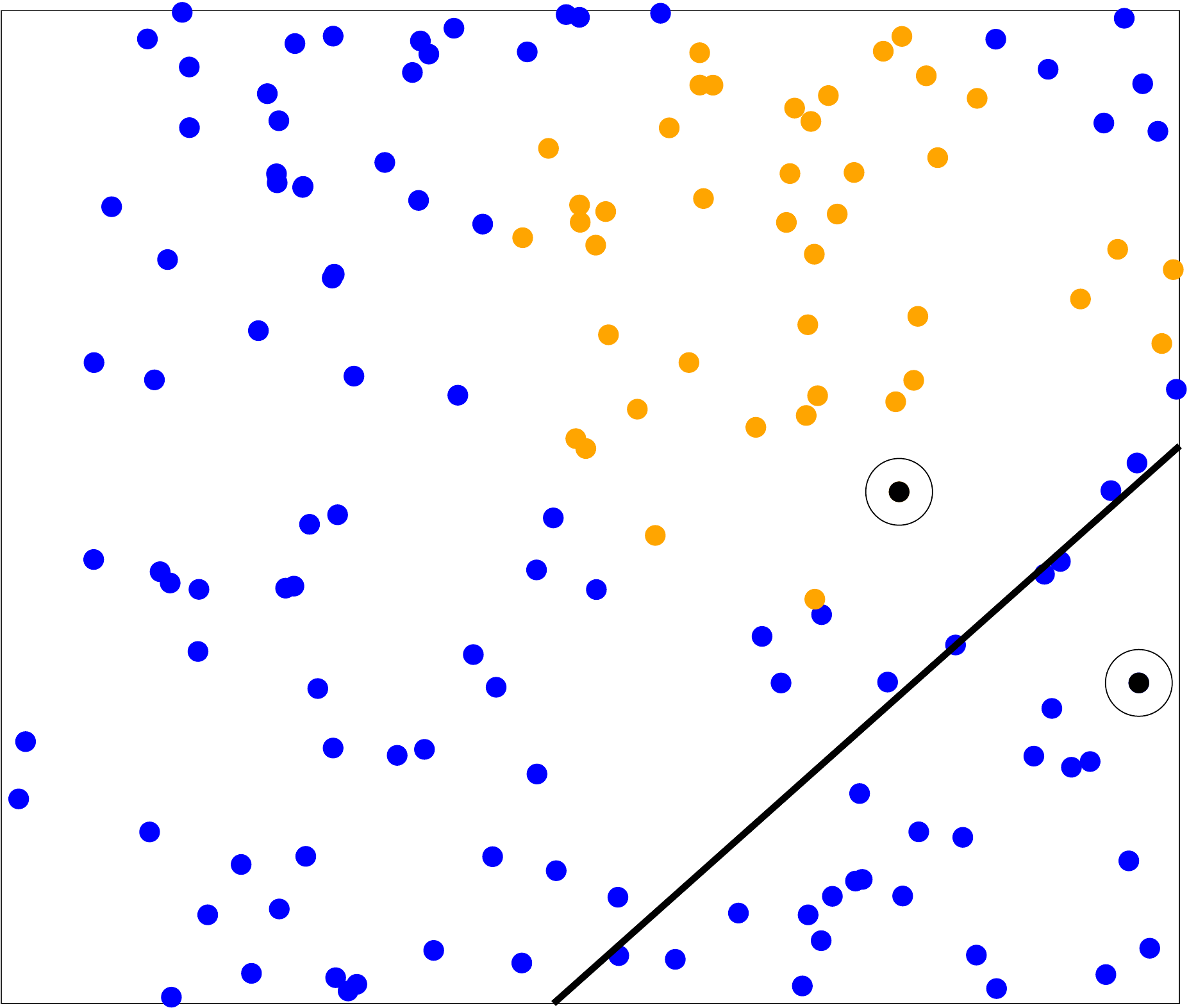}
    \end{subfigure}\hfill
    \begin{subfigure}
    \centering
    \raisebox{1.2cm}{\includegraphics[width=.10\linewidth]{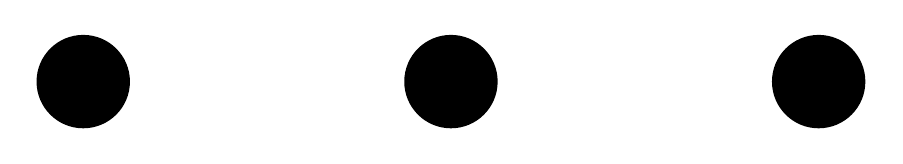}}
    \end{subfigure}\hfill
    \begin{subfigure}
    \centering
    \includegraphics[width=.185\linewidth]{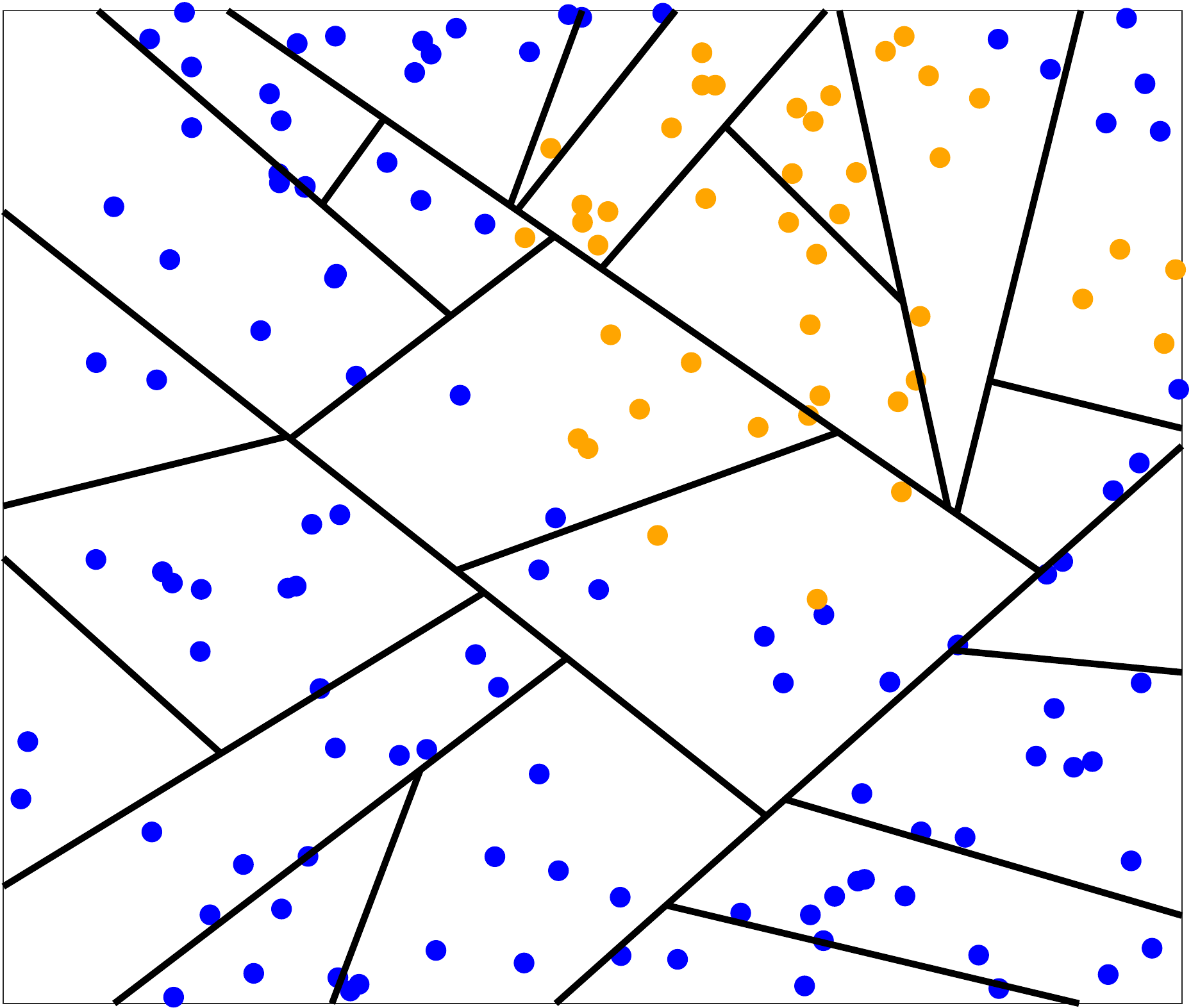}
    \end{subfigure}\hfill
    \caption{\label{fig:compr-tree-construction}Construction of the comparison tree, illustrated in the Euclidean setting. (i) The current cell contains points with two different labels. (ii) Two pivot points with opposite labels are chosen randomly from all sample points in the current cell (circled black dots). (iii) The current cell is split according to whether points are closer to the one or the other pivot; in the Euclidean setting  this corresponds to a hyperplane split. (iv) Result after recursive application of this principle with final leaf size $n_0=10$. }
    \vskip -0.2in
\end{figure*}

Random forests, first introduced in~\citet{breiman2001random}, are one of the most popular algorithms for classification and regression in Euclidean spaces. In a comprehensive study on more than $100$ classification tasks, random forests show the best performance among many other general purpose methods~\citep{fernandez2014we}. 
However, standard random forests heavily rely on the vector space representation of the underlying data points, which is not available in a comparison-based framework. 
Instead, we propose a comparison-based random forest algorithm for classification and regression tasks. The main ingredient is the comparison tree, which only uses of triplet comparisons and does not rely on Euclidean representation or distances between items. 

Let us recap the {\bf CART random forest:}
The input consists of a labeled set $D_n = \{ (x_1,y_1),(x_2,y_2),\ldots,(x_n,y_n) \} \subset \Reals^d \times \Reals$. To build an individual tree, we first draw a random subsample $D_{s}$ of $a_n$ points from $D_n$. Second, we select a random subset $Dim_{s}$ of size $\texttt{mtry}$ of all possible dimensions $\{1,2,\ldots,d\}$.
The tree is then built based on recursive, axis-aligned splits along a dimension randomly chosen from $Dim_{s}$. 
The exact splitting point along this direction is determined via the CART criterion, which also involves the labels of the subset $D_{s}$ of points (see \citet{biau2016random} for details). The tree is grown until each cell contains at most~$n_0$ points---these cells then correspond to the leaf nodes of the tree. To estimate a regression function $m(x)$, each individual tree routes the query point to the appropriate leaf and outputs the average response over all points in this leaf. 
The random forest aggregates~$M$ such trees. 
Let us denote the prediction of tree~$i$ at point~$x$ by $m_i(x,\Theta_i,D_n)$, where $\Theta_i$ encodes the randomness in the tree construction.
Then the final forest estimation at~$x$ is the average result over all trees  (for classification, the average is replaced by a majority vote):
\[
m_{M,n}(x;\left(\Theta_i\right)_{1\leq i\leq M},D_n) = \frac{1}{M}\sum_{i=1}^{M}{m_i(x,\Theta_i,D_n)}
\, .
\]
The general consensus in the literature is that CART forests are surprisingly robust to parameter choices. 
Consequently, people use explicit rules of thumb, for example to set  $\texttt{mtry} =\ceil {d/3}$, and $n_0=5$ (resp. $n_0=1$) for regression (resp. classification) tasks.

We now suggest to replace CART trees by comparison trees, leading to {\bf comparison-based random forests (CompRF)}. Comparison trees have originally been designed to find nearest neighbors by recursively splitting the search space into smaller subspaces. Inspired by the CART criterion, we propose a supervised variant of the comparison tree, which we refer to as ``supervised comparison tree.''

\begin{algorithm}
\caption{$CompTree(S,n_0)$: \\\textbf{Supervised comparison tree construction}}
\label{alg:CompTree}
\begin{algorithmic}[1]
\REQUIRE Labeled data $S$ and maximum leaf size $n_0$
\ENSURE Comparison tree $T$
\STATE $T.root \gets S$
\IF{$|S| > n_0$}
\STATE Sample distinct $(x_1,y_1),(x_2,y_2) \in S$ s.t. $y_1\neq y_2$\\
 (if all points have the same label choose randomly)
\STATE $S_1 \gets \{(x,y) \in S : \delta(x,x_1) \leq \delta(x,x_2)\}$
\STATE $T.leftpivot \gets x_1,$~~$T.rightpivot \gets x_2$
\STATE $T.leftchild \gets CompTree(S_1,n_0)$
\STATE $T.rightchild \gets CompTree(S \backslash S_1,n_0)$
\ENDIF
\STATE \textbf{Return} $T$
\end{algorithmic}
\end{algorithm}

{\bf For classification}, the supervised comparison tree construction for a labeled set $S \subset \Xc \times \{0,1\}$ is as follows (see Algorithm~\ref{alg:CompTree} and Figure~\ref{fig:compr-tree-construction}): we randomly choose two pivot points $x_1$ and $x_2$ with different labels~$y_1$ and~$y_2$ 
among the points in~$S$ (the case where all the points in~$S$ have the same label is trivial). 
For every remaining point $(x,y) \in S$, we request the triplet comparison ``$\delta(x,x_1)< \delta(x,x_2)$.'' 
The answer to this query determines the relative position of~$x$ with respect to the generalized hyperplane separating~$x_1$ and~$x_2$. 
We assign the points closer to~$x_1$ to the first child node of~$S$ and the points closer to~$x_2$ to the other one.  
We now recurse the algorithm on the child nodes until less than~$n_0$ points remain in every leaf node of the tree. 

The supervised pivot selection is analogous to the CART criterion. However, instead of a costly optimization over the choice of split, it only requires to choose pivots with different labels. In Section~\ref{subsec:EucExps}, we empirically show that the supervised split procedure leads to a better performance than the CART forests for classification tasks. 

{\bf For regression}, it is not obvious how the pivot selection should depend on the output values. Here we use an unsupervised version of the forest (unsupervised CompRF): we choose the pivots $x_1,x_2$ without considering $y_1,y_2$.

The final comparison-based forest consists of $M$ independently constructed comparison trees. To assign a label to a query point, we traverse every tree to a leaf node, then we aggregate all the items in the leaf nodes of $M$ trees to estimate the label of the query item. For classification, the final label is the majority vote over the labels of the accumulated set (in the multiclass case we use a one vs. one approach). For regression we use the mean output value. 

\begin{algorithm}[t]
\caption{$CompRF(D_n,q,M,n_0,r)$: \\\textbf{CompRF prediction at query $q$}}
\begin{algorithmic}[1]
\REQUIRE Labeled dataset $D_n \subset \Xc \times \{0,1\}$, query $q \in \Xc$, leaf size $n_0$, trees $M$ and subsampling ratio $r$.
\ENSURE $y_q=$  label prediction for $q$
\STATE Set $C= \emptyset$ as the list of predictor items
\FOR {j=1,\ldots,M}
\STATE Take a random subsample $D_s \subset D_n$, s.t., $\frac{\vert D_s\vert}{\vert D_n \vert} = r$ 
\STATE $T_j \gets CompTree(D_s,n_0)$
\STATE Given $q$, traverse the tree $T_j$ to the leaf node $N_j$
\STATE $C \gets C \cup N_j$
\ENDFOR
\STATE \textbf{Return} MajorityVote($\{y\vert (x,y) \in C\}$)
\end{algorithmic}
\label{alg:CompRF}
\end{algorithm}

{\bf Intuitive comparison:} The general understanding is that the efficiency of CART random forests is due to:  (1) the randomness due to subsampling of dimensions and data points  \citep{breiman1996bagging}; (2) the CART splitting criterion that exploits the label information already in the tree construction \citep{breiman1984classification}. 
A weakness of CART splits is that they are necessarily axis-aligned, and thus not well-adapted to the geometry of the data. 

In comparison trees, randomness is involved in the tree construction as well. But once a splitting direction has been determined by choosing the pivot points, the exact splitting point along this direction cannot be influenced any more, due to the lack of a vector space representation. 
On the other hand, the comparison tree splits are well adapted to the data geometry by construction, giving some advantage to the comparison trees. 

All in all, the comparison-based forest is a promising candidate with slightly different strengths and weaknesses than CART forest. Our empirical comparison in Section~\ref{subsec:EucExps} reveals that it performs surprisingly well and can even outperform CART forests in certain settings. 

\section{Theoretical Analysis}
\label{sec:analysis}

Despite their intensive use in practice, theoretical questions regarding the consistency of the original procedure of~\citet{breiman2001random} are still under investigation.
Most of the research focuses on simplified models in which the construction of the forest does not depend on the training set at all \citep{Bia:2012}, or only via the $x_i$s but not the $y_i$s  \citep{Bia_Dev_Lug:2008,Ish_Kog:2010,Den_Mat_Fre:2013}.
Recent efforts nearly closed this gap, notably \citet{Sco_Bia_Ver:2015}, where it is shown that the original algorithm is consistent in the context of additive regression models and under suitable assumptions.
However, there is no previous work on the consistency of random forests constructed only with triplet comparisons.  

As a first step in this direction, we investigate the consistency of individual comparison trees, which is the first building block in the study of random forests consistency.
As it is common in the theoretical literature on random forests, we consider a slightly modified version of the comparison tree. We assume that the pivot points are not randomly drawn from the underlying sample but according to the true distribution of the data. 
In this setting, we show that, when the number of observations grows to infinity, 
(i) the diameter of the cells converges to zero in probability, and 
(ii) each cell contains an arbitrarily large number of observations.
Using a result of~\citet{Dev_Gyo_Lug:1996}, we deduce that the associated classifier is consistent.
The challenging part of the proof is to obtain control over the diameter of the cells. 
Intuitively, as in \citet[Lemma~12]{dasgupta2008random}, it suffices to show that each cut has a larger probability to decrease the diameter of the current cell than that of leaving it unchanged.
To prove this in our case is very challenging since both the position and the decrease in diameter caused by the next cut depend on the \emph{geometry} of the cell.

\subsection{Continuous Comparison Tree}
\label{sec:analysis-notations}

As it is the case for most theoretical results on random forests, we carry out our analysis in a Euclidean setting (however, the comparison-forest only has indirect access to the Euclidean metric via triplet queries). 
We assume that the input space is $\Inputspace =[0,1]^{\Dim}$ with distance~$\Dist$ given by the Euclidean norm, that is, $\dist{x}{y}=\norm{x-y}$.
Let~$X$ be a random variable with support included in $[0,1]^{\Dim}$.
We assume that the observations $X_1,\ldots,X_n\in [0,1]^{\Dim}$ are drawn independently according to the distribution of~$X$.
We make the following assumptions:  
\begin{assumption}
\label{assump:density}
The random variable $X\in [0,1]^{\Dim}$ has density~$\Density$ with respect to the Lebesgue measure on $[0,1]^{\Dim}$.
Additionally, $\Density$ is finite and bounded away from~$0$. 
\end{assumption}

For any $x,y\in\Reals^\Dim$, let us define 
\[
\Delta(x,y) \Defeq \normalcondset{z\in\Reals^\Dim}{\dist{x}{z}=\dist{y}{z}}
\, .
\]
In the Euclidean setting, $\Delta(x,y)$ is a hyperplane that separates $\Reals^\Dim$ in two half-spaces.
We call $H_x$ (resp.~$H_y$) the open half-space containing~$x$ (resp.~$y$).
The set~$S_1$ in Algorithm~\ref{alg:CompTree} corresponds to $S\cap H_{x_1}$.

We can now define the continuous comparison tree.
\begin{definition}[{\bf Continuous comparison tree}]
A \emph{continuous comparison tree} is a random infinite binary tree~$T^0$ obtained \emph{via} the following iterative construction:
\begin{itemize}[topsep=0pt, partopsep=0pt]
	\item 
	The root of $T^0$ is $[0,1]^\Dim$;
	
	\item
	Assuming that level~$\ell$ of $T^0$ has been built already, then level $\ell+1$ is built as follows: 
	For every cell~$C$ at height~$\ell$, draw $X_1,X_2\in C$ independently according to the distribution of~$X$ restricted to~$C$.
	The children of~$C$ are defined as the closure of $C\cap H_{X_1}$ and $C\cap H_{X_2}$.
\end{itemize}
For any sequence $\left(p_n\right)_{n\geq 0}$, a \emph{truncated}, continuous comparison tree $T^0(p_n)$ consists of the first $\floor{p_n}$ levels of~$T^0$.
\end{definition}

From a mathematical point of view, the continuous tree has a number of advantages. (i) Its construction does not depend on the responses $Y_1,\ldots,Y_n$.  Such a simplification is quite common because  data-dependent random tree structures are notoriously difficult to analyze~\citep{Bia_Dev_Lug:2008}. (ii) Its construction is formally  independent of the finite set of data points, but ``close in spirit'':  
Rather than sampling the pivots among the data points in a cell, pivots are  independently sampled according to the underlying distribution.  
Whenever a cell contains a large number of sample points, both distributions are close, but they may drift apart when the diameter of the cells go to~$0$. 
(iii) In the continuous comparison tree, we stop splitting cells at height $\floor{p_n}$, whereas in the discrete setting we stop if there is less than~$n_0$ observations in the current cell. 
As a consequence, $T^0(p_n)$ is a perfect binary tree: each interior node has exactly~$2$ children. 
This is typically not the case for comparison trees. 

\subsection{Consistency}
\label{sec:analysis-main-results}

To each realization of $T^0(p_n)$ is associated a partition of $[0,1]^\Dim$ into disjoint cells 
$
A_{1,n},A_{2,n},\ldots,A_{2^{p_n},n}
\, .
$
For any $x\in [0,1]^{\Dim}$, let $A(x)$ be the cell of $T^0(p_n)$ containing~$x$.
Let us assume that the responses $\left(Y_i\right)_{1\leq i\leq n}$ are binary labels.
We consider the classifier defined by majority vote in each cell, that is,
\[
g_n(x) \Defeq
\begin{cases}
1 \enspace\text{if}\enspace \sum_{X_i\in A(x)}\indic{Y_i=1} \geq \sum_{X_i\in A(x)} \indic{Y_i=0} \\
0 \enspace \text{otherwise.}
\end{cases}
\]
Define 
$
L_n \Defeq \condproba{g_n(X)\neq Y}{D_n}
.
$
Following \citet{Dev_Gyo_Lug:1996}, we say that the classifier $g_n$ is \emph{consistent} if
\[
\expec{L_n} = \proba{g_n(X)\neq Y} \xrightarrow[n\to +\infty]{} L^{\star}
\, ,
\]
where $L^{\star}$ is the Bayes error probability.
Our main result is the consistency of the classifier associated with the continuous comparison tree truncated to a logarithmic height.

\begin{theorem}[{\bf Consistency of comparison-based trees}]
\label{th:consistency}
Under  Assumption~\ref{assump:density}, the classifier associated to the continuous, truncated tree $T^0(\alpha \log n)$ is consistent for any constant $0<\alpha < 1/\log 2$.
\end{theorem}

In particular, since each individual tree is consistent, a random forest with base tree $T^0(p_n)$ is also consistent. 
Theorem~\ref{th:consistency} is a first step towards explaining why comparison-based trees perform well without having access to the representation of the points. 
Also note that, even though the continuous tree is a simplified version of the discrete tree, they are quite similar and share all important characteristics. 
In particular, they roughly have the same depth---with high probability, the comparison tree has logarithmic depth~\citep[Theorem~1]{haghiri2017comparison}.

\begin{table*}[t]
     \caption{\label{tab:Classification}Average and standard deviation of classification error for the CompRF vs. other methods. The first three rows describe datasets.}
    \vskip 0.15in
    \ra{1.0}
    \centering
     \begin{tabular}{@{}ccccc@{}} \toprule
            & MNIST & Gisette & UCIHAR & Isolet \\ \midrule
          Dataset Size & 70000 & 7000 & 10229 & 6238 \\
          Variables & 728 & 5000 & 561 & 617 \\ 
          Classes & 10 & 2 & 5 & 26 \\ \midrule
         
          KNN & 2.91 & 3.50 & 12.15 & 8.27 \\ 
         
          CART RF & 2.90 ($\pm$ 0.05) & 3.04 ($\pm$ 0.26) & 7.47 ($\pm$ 0.32) & 5.48 ($\pm$ 0.27) \\ 
         
          CompRF unsupervised & 4.21 ($\pm$ 0.05) & 3.28 ($\pm$ 0.19) & 8.70 ($\pm$ 0.32) & 6.65 ($\pm$ 0.14)\\ 
         
          CompRF supervised & \textbf{2.50} ($\pm$ 0.05) & \textbf{2.48} ($\pm$ 0.13) & \textbf{6.54} ($\pm$ 0.11) & \textbf{4.43} ($\pm$ 0.26) \\ \bottomrule
     \end{tabular}
     \vskip -0.1in
\end{table*}

\begin{figure*}[ht]
    \vskip 0.2in
    \centering
    \begin{subfigure}
    \centering
    \includegraphics[width=.22\linewidth]{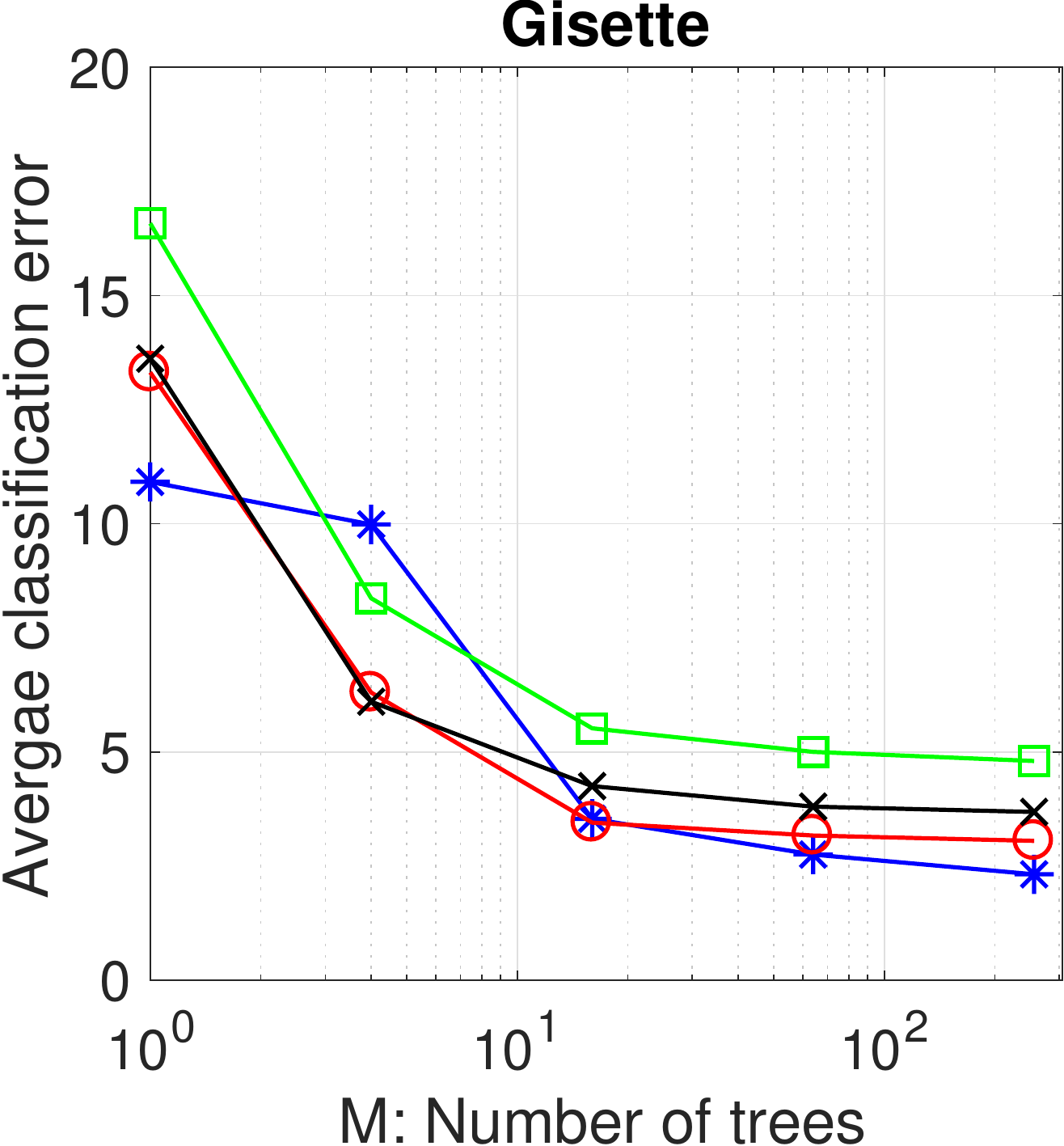}
    \end{subfigure}
    \begin{subfigure}
    \centering
    \includegraphics[width=.22\linewidth]{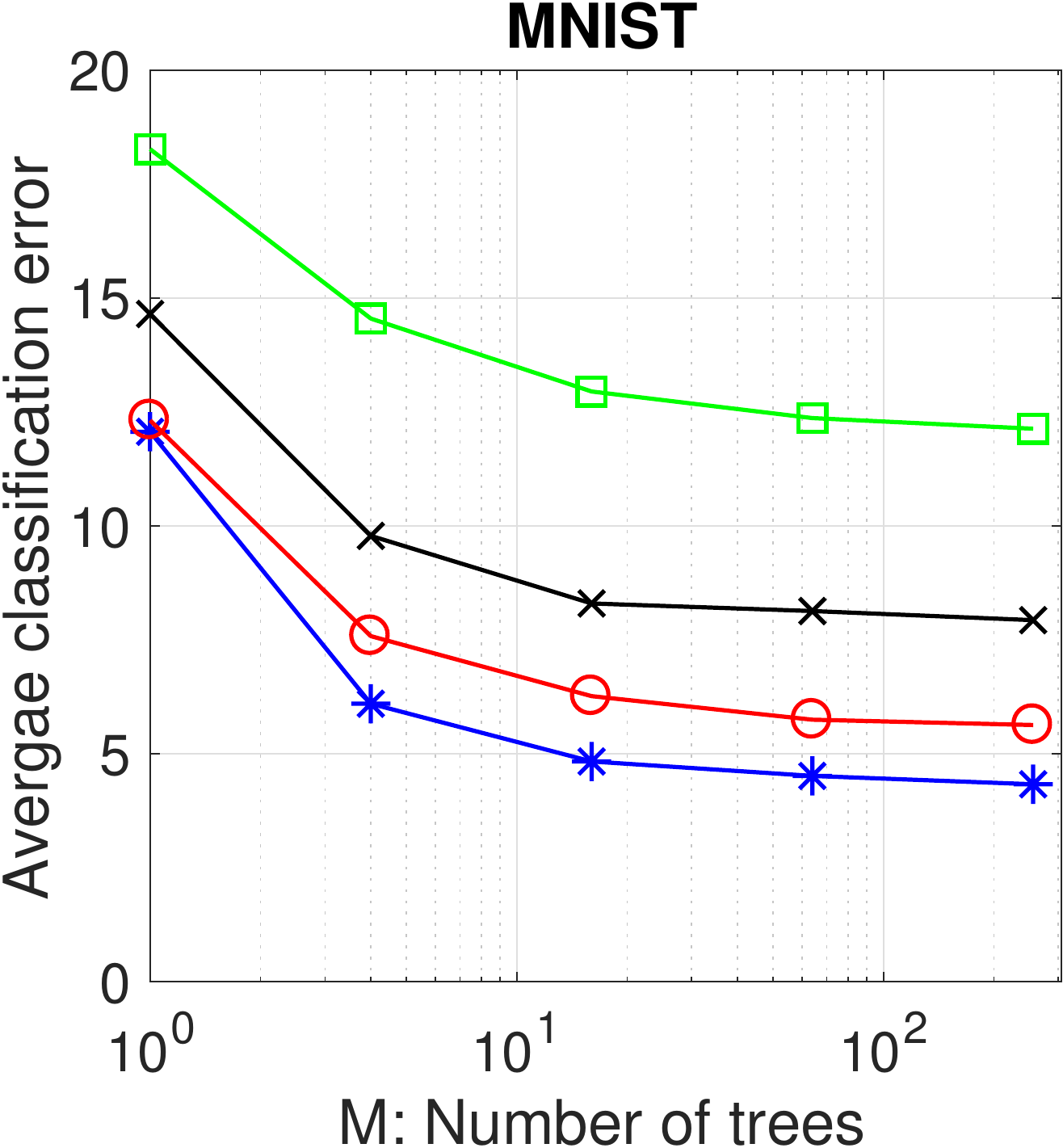}
    \end{subfigure}
    \begin{subfigure}
    \centering
    \includegraphics[width=.22\linewidth]{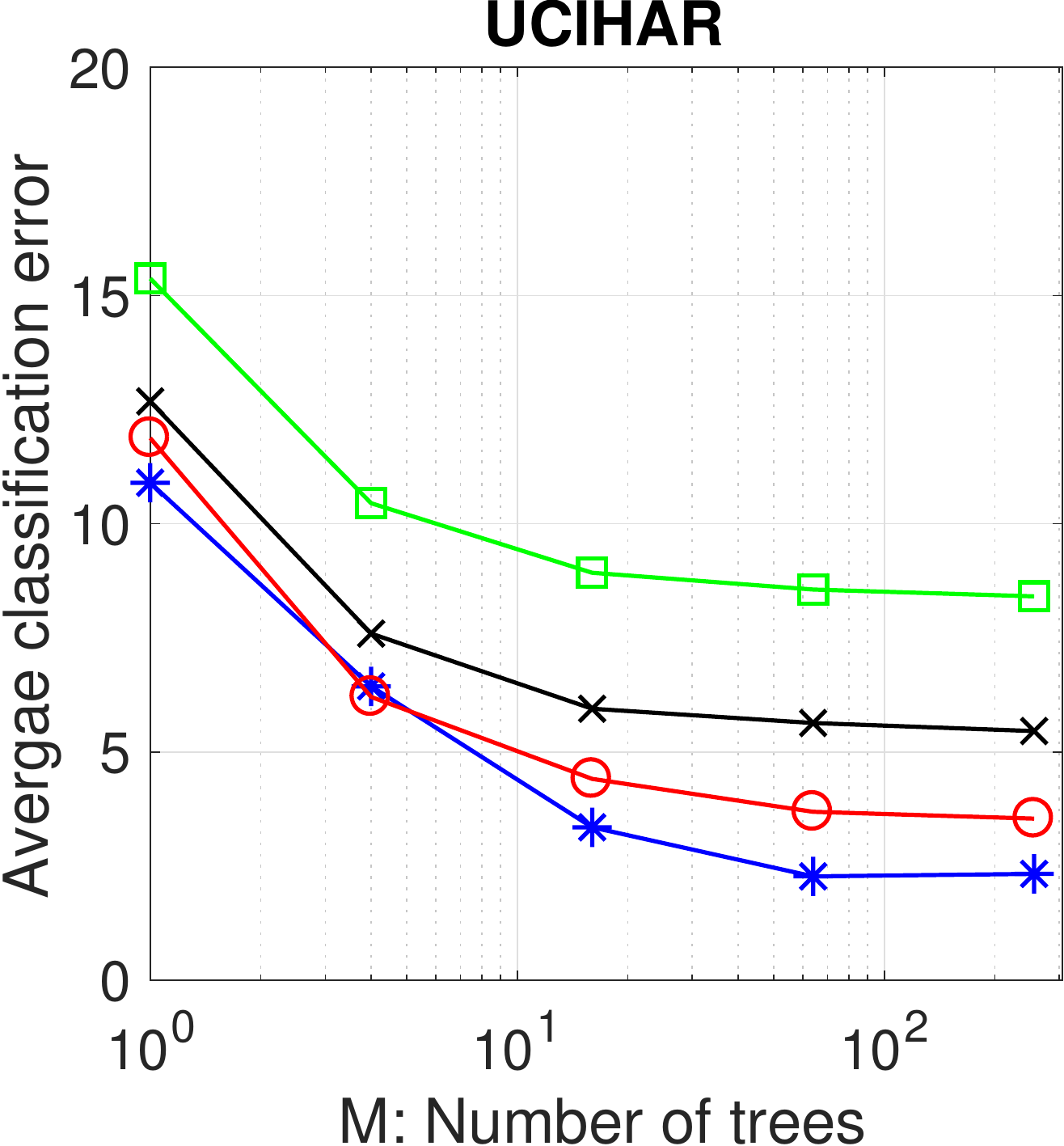}
    \end{subfigure}
    \begin{subfigure}
    \centering
    \includegraphics[width=.22\linewidth]{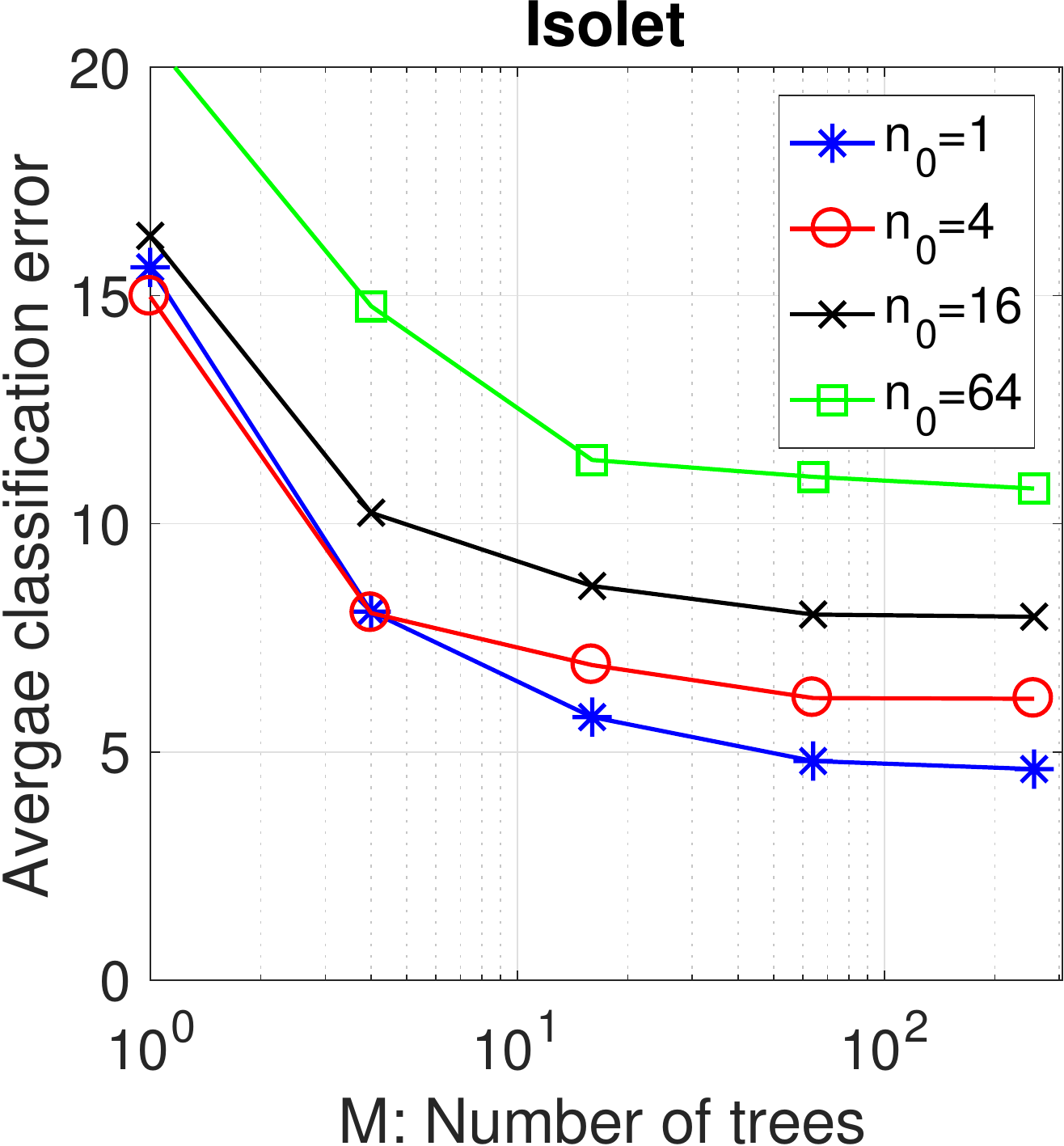}
    \end{subfigure}
    \caption{\label{fig:ClsExp1}Average classification error of the CompRF algorithm on classification datasets. X-Axis shows the number of trees used in the forest. The title denotes the dataset and each curve corresponds to a fixed value of $n_0$. }
    \vskip -0.2in
\end{figure*}

\subsection{Sketch of the Proof}
\label{sec:analysis-sketch-proof}

Since the construction of $T^0(p_n)$ does not depend on the labels, we can use Theorem~6.1 of~\citet{Dev_Gyo_Lug:1996}. 
It gives sufficient conditions for classification rules based on space partitioning to be consistent. 
In particular, we have to show that the partition satisfies two properties: 
first, the leaf cells should be small enough, so that local changes of the distribution can be detected; 
second, the leaf cells should contain a sufficiently large number of points so that averaging among the labels makes sense.
More precisely, we have to show that (1) $\diam{A(X)} \to 0$ in probability, where $\diam{A} \Defeq \sup_{x,y\in A}\dist{x}{y}$ is the diameter of~$A$, and (2) $N(X)\to \infty$ in probability, where $N(X)$ is the number of sample points in the cell containing~$X$.
The second point is simple, because  it is sufficient to show that the number of cells in the partition associated to $T^0(\alpha \log n)$ is $\littleo{n}$ (according to Lemma~20.1 in \citet{Dev_Gyo_Lug:1996} and the remark that follows). 
Proving (1) is much more challenging.
A sketch of the proof of (1) follows---note that the complete version of the proof of Theorem~\ref{th:consistency} can be found in the supplementary material.

The critical part in proving (1) is to show that, for any cell of the continuous comparison tree, the diameter of its descendants at least~$k$ levels below is halved with high probability. 
More precisely, the following proposition shows that this probability is lower bounded by $1-\delta$, where $\delta$ is exponentially decreasing in~$k$.

\begin{proposition}[{\bf Diameter control}]
\label{prop:diameter-control}
Let~$C$ be a cell of $T^0(X)$ such that $\diam{C}\leq D$.
Then, under Assumption~\ref{assump:density}, the probability that there exists a descendant of~$C$ which is more than~$k$ levels below and yet has diameter greater than $D/2$ is at most $\delta = c\gamma^k$, where $c>0$ and $\gamma\in(0,1)$ are constants depending only on the dimension~$\Dim$ and the density~$\Density$. 
\end{proposition}

The proof of Proposition~\ref{prop:diameter-control} amounts to showing that the probability of decreasing the diameter of any given cell is higher than the probability of keeping it unchanged---see the supplementary material.

Assuming Proposition~\ref{prop:diameter-control}, the rest of the proof goes as follows.
Let us set $\epsilon \in(0,1)$.
We are going to show that 
\[
\proba{\diam{A(X)}>\epsilon} \longrightarrow 0 \quad \text{when}\quad n\to +\infty
\, .
\]
Let $\Gamma$ be the path in $T^0(\alpha\log n)$ that goes from the root to the leaf~$A$ of maximum diameter.
This path has length $\floor{p_n}$ according to the definition of $T^0(\alpha\log n)$.
The root, which consists of the set $[0,1]^d$, has diameter $\sqrt{\Dim}$. 
This means that we need to divide the diameter of this cell $\pi=\ceil{\log_2 \sqrt{\Dim}/\epsilon}$ times to obtain cells with diameter smaller than~$\epsilon$.
Let us set $k=\floor{p_n/\pi}$ and pick cells $\left(C^{(j)}\right)_{0\leq j\leq \pi}$ along $\Gamma$ such that $C^{(0)}=[0,1]^{\Dim}$, $C^{(\pi)}=A$, and such that there are more than~$k$ levels between $C^{(j)}$ and $C^{(j+1)}$.
Then we can prove that $\proba{\diam{A} > \epsilon}$ is smaller than
\[
\sum_{j=1}^{\pi} \condproba{\diam{C^{(j)}}>\dfrac{\sqrt{\Dim}}{2^{j}}}{\diam{C^{(j-1)}}\leq \frac{\sqrt{\Dim}}{2^{j-1}}}
\, .
\]
Furthermore, according to Prop.~\ref{prop:diameter-control}, the quantity in the last expression is upper bounded by $\pi c\gamma^k$.
Since $k=\bigo{\log n}$ and $\gamma\in (0,1)$, we can conclude. \qed

\begin{table*}[t]
    \caption{\label{tab:reg_results}Average and standard deviation of the RMSE for the CompRF vs. CART regression forest. 
    }
    \vskip 0.15in
    \ra{1.0}
    \centering
    \begin{tabular}{@{}ccccc@{}} \toprule
          & ONP & Boston & ForestFire & WhiteWine \\ \midrule
         Dataset Size & 39644 & 506 & 517 & 4898 \\
         Variables & 58 & 13 & 12& 11 \\ \midrule
         CART RF & \textbf{1.04} ($\pm$ 0.50) $\cdot 10^4$ & \textbf{3.02} ($\pm$ 0.95) & \textbf{45.32} ($\pm$ 4.89) & \textbf{59.00} ($\pm$ 2.94)$\cdot 10^{-2}$\\
         CompRF & 1.05 ($\pm$ 0.50) $\cdot 10^4$& 6.16 ($\pm$ 1.00) & 45.37 ($\pm$ 4.69) & 72.46 ($\pm$ 3.16) $\cdot 10^{-2}$\\ \bottomrule
    \end{tabular}
    \vskip -0.10in
\end{table*}
\begin{figure*}[t]
    \vskip 0.2in
    \centering
    \begin{subfigure}
    \centering
    \includegraphics[width=.22\linewidth]{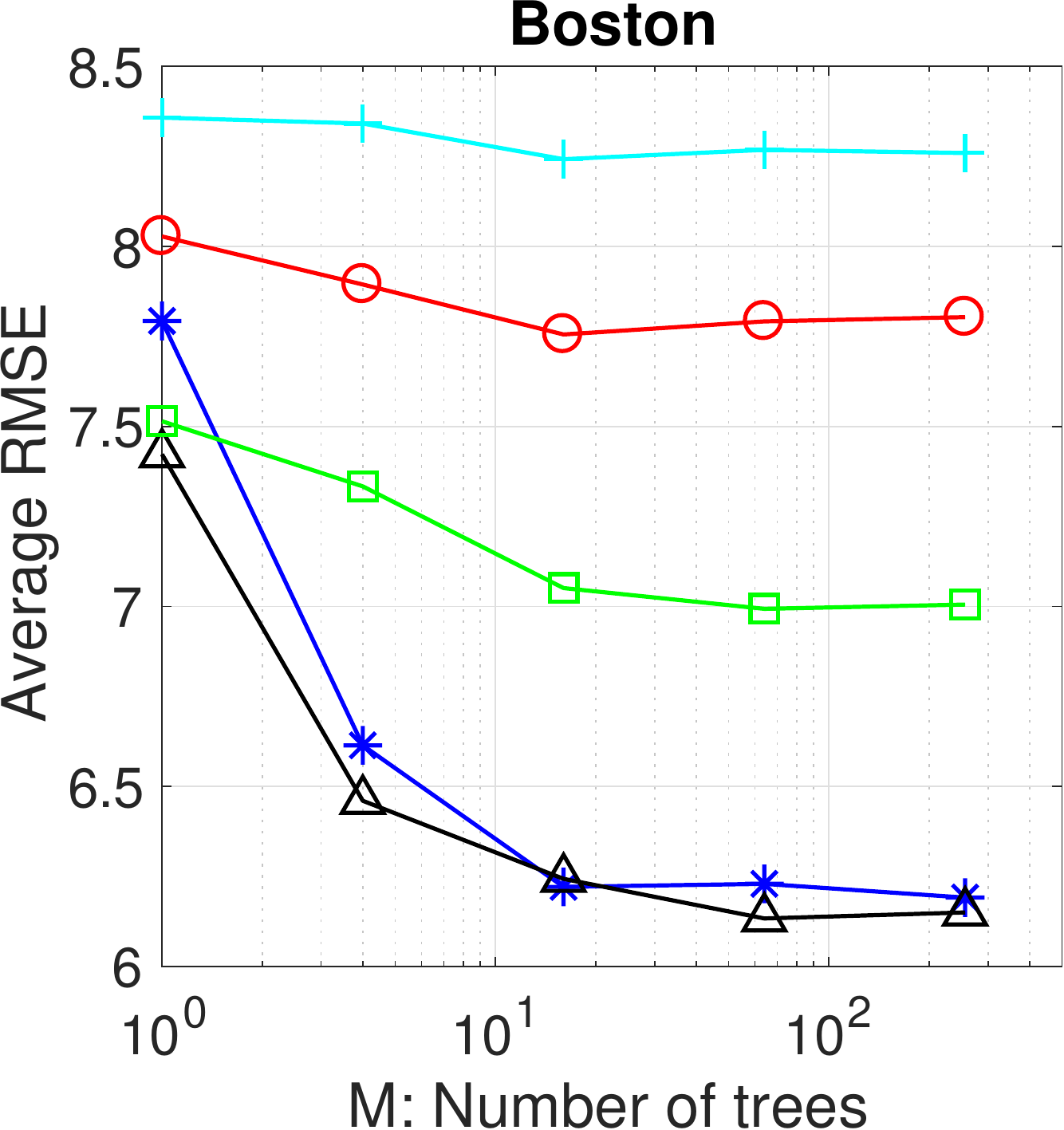}
    \end{subfigure}
    \begin{subfigure}
    \centering
    \includegraphics[width=.225\linewidth]{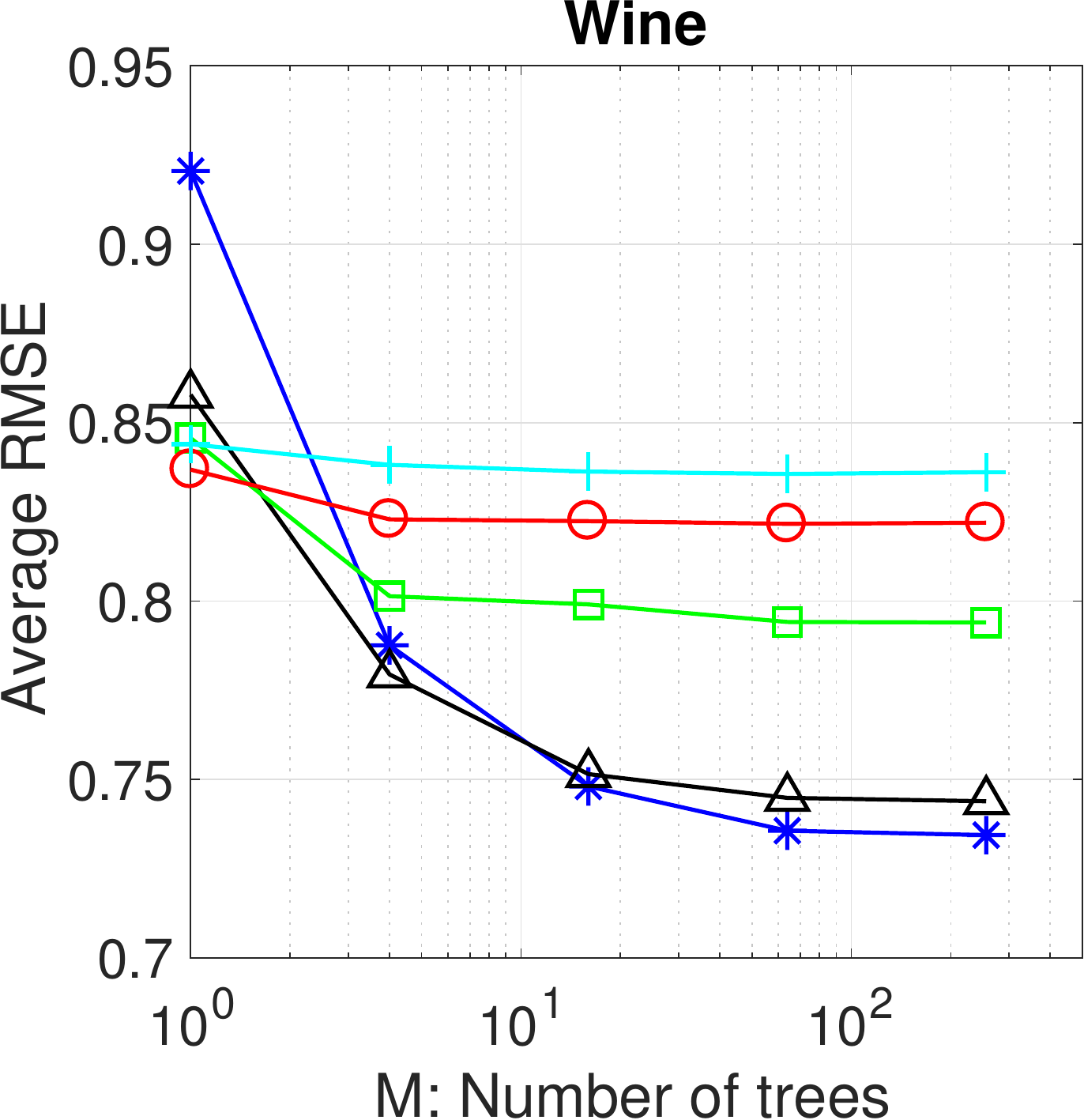}
    \end{subfigure}
    \begin{subfigure}
    \centering
    \includegraphics[width=.215\linewidth]{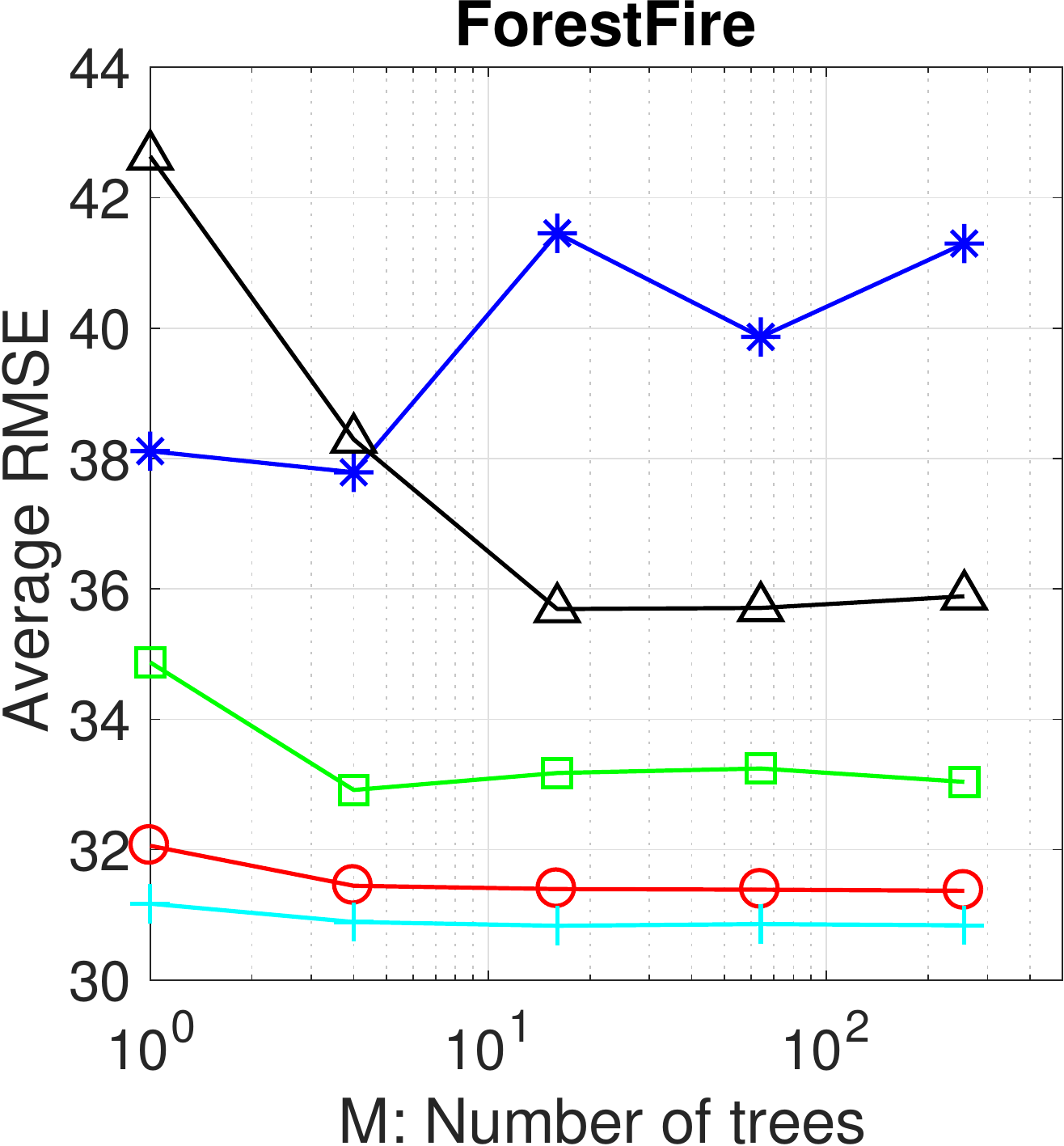}
    \end{subfigure}
    \begin{subfigure}
    \centering
    \includegraphics[width=.22\linewidth]{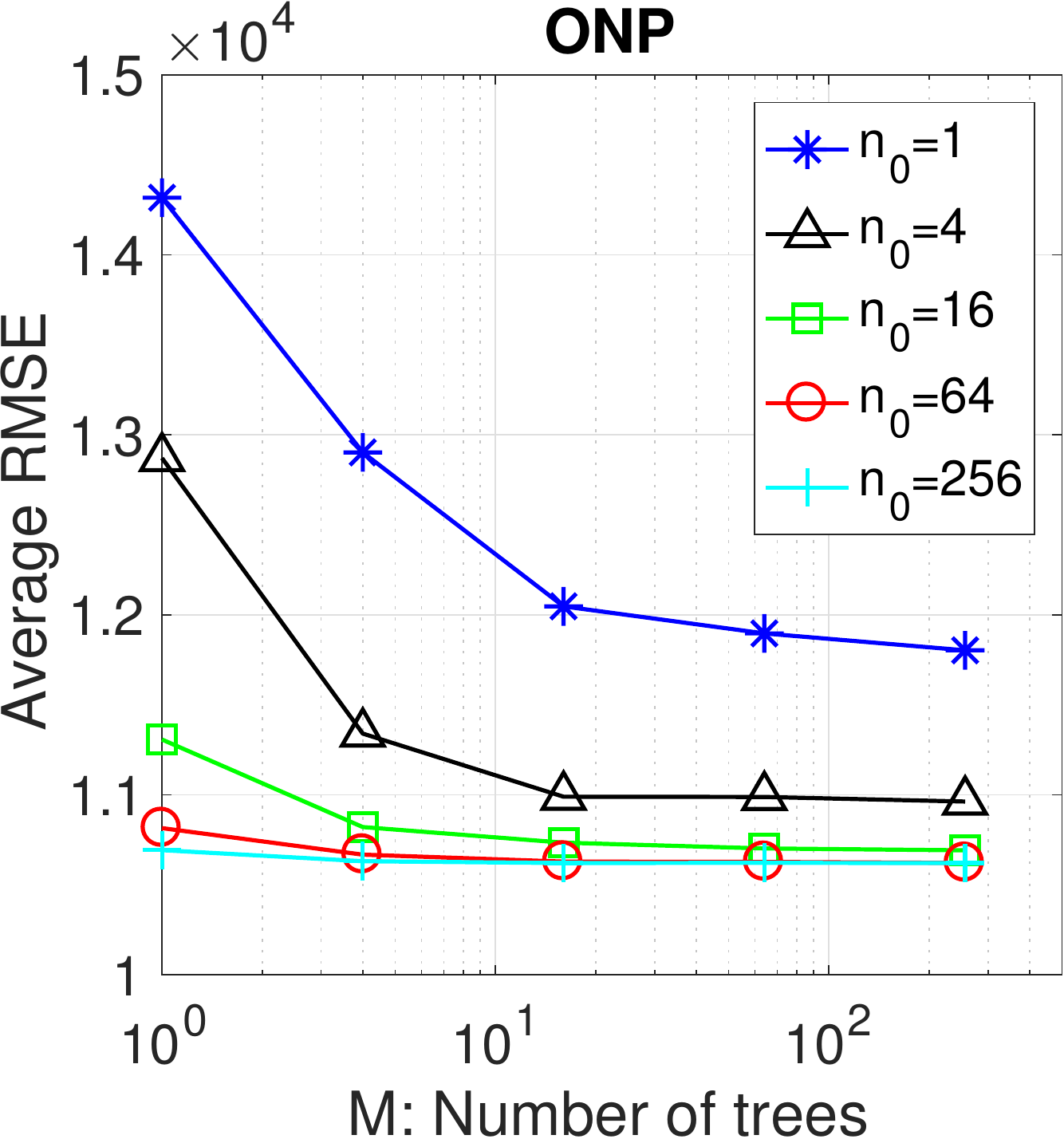}
    \end{subfigure}
    \caption{\label{fig:RegExp1}Average RMSE of the CompRF algorithm on regression datasets. X-Axis shows the number of trees used in the forest. The title denotes dataset and each curve corresponds to a fixed value of $n_0$.}
    \vskip -0.2in
\end{figure*}

\section{Experiments}
\label{sec:experiments}

In this section, we first examine comparison-based forests in the Euclidean setting. Secondly, we apply the CompRF method to non-Euclidean datasets with a general metric available. Finally we run experiments in the setting where we are only given triplet comparisons. 

\subsection{Euclidean Setting}
\label{subsec:EucExps}

Here we examine the performance of CompRF on classification and regression tasks in the Euclidean setting, and compare it against CART random forests as well as the KNN  classifier as a baseline. As distance function for KNN and CompRF we use the standard Euclidean distance. 
Since the CompRF only has access to distance comparisons, the amount of information it uses is considerably lower than the information available to the CART forest. Hence, the goal of this experiment is not to show that comparison-based random forests can perform better, but rather to find out whether the performance is still acceptable. 

To emphasize the role of supervised pivot selection, we report the performance of the unsupervised CompRF algorithm in classification tasks as well. 
The tree structure in the unsupervised CompRF chooses the pivot points uniformly at random without considering the labels.

For the sake of simplicity, we do not perform subsampling when building the CompRF trees. We report some experiments concerning the role of subsampling in Section~3.2 of supplementary material. All other parameters of CompRF are adjusted by cross-validation. 

\subsubsection{Classification} 

We use four classification datasets. MNIST~\citep{mnist} and Gisette are handwritten digit datasets. Isolet and UCIHAR are speech recognition and human activity recognition datasets respectively~\citep{Lichman:2013}. Details of the datasets are shown in the first three rows of Table~\ref{tab:Classification}. 

\textbf{Parameters of CompRF:} We examine the behaviour of the CompRF with respect to the choice of the leaf size $n_0$ and the number of trees $M$. 
We perform $10$-fold cross-validation over $n_0 \in \{1,4,16,64\}$ and $M \in \{1,4,16,64,256\}$. 
In Figure~\ref{fig:ClsExp1} we report the resulting cross validation error. 
Similar to the recommendation for CART forests~\citep{biau2016random}, we achieve the best performance when the leaf size is small, that is  $(n_0=1)$.  
Moreover, there is no significant improvement for $M$ greater than $100$.

\textbf{Comparison between CompRF, CART and KNN:} 
Table~\ref{tab:Classification} shows the average and standard deviation of classification error for $10$ independent runs of CompRF, CART forest and KNN. Training and test sets are given in the respective datasets. The parameters $n_0$ and $M$ of CompRF and CART, and $k$ of KNN are chosen by $10$-fold cross validation on the training set. 
Note that KNN is not randomized, thus there is no standard deviation to report.

\begin{table*}[ht]
    \caption{\label{tab:kernelbased}Average and standard deviation of the classification error for the CompRF in comparison with kernelSVM on graph datasets with two graph kernels: WL-subtree and WL-edge. 
    }
    \vskip 0.15in
    \ra{1.0}
    \centering
    \begin{tabular}{@{}ccccc@{}} \toprule
          & MUTAG & ENZYMES & NCI1 & NCI109  \\ \midrule
         Train Size & 188 & 600 & 4110 & 4127  \\ 
         Classes & 2 & 6 & 2 & 2 \\ \midrule
         
         \multicolumn{5}{c}{\textbf{WL-subtree kernel}} \\ \midrule
         
         Kernel SVM & 17.77 ($\pm$ 7.31) & 47.16 ($\pm$ 5.72) & \textbf{15.96} ($\pm$ 1.56) & \textbf{15.55} ($\pm$ 1.40) \\ 
        
        KNN & 14.00 ($\pm$ 8.78) & 48.17 ($\pm$ 4.48) & 18.13 ($\pm$ 2.27) & 18.74 ($\pm$ 1.97) \\ 
        
         CompRF unsupervised & 14.44 ($\pm$ 7.94) & \textbf{39.33} ($\pm$ 6.49) & 17.96 ($\pm$ 1.85) & 19.10 ($\pm$ 2.22) \\ 
         
         CompRF supervised & \textbf{13.89} ($\pm$ 7.97) & 39.83 ($\pm$ 5.00) & 17.35 ($\pm$ 1.98) & 18.71 ($\pm$ 2.61) \\ \midrule
         
         \multicolumn{5}{c}{\textbf{WL-edge kernel}} \\ \midrule
         
         Kernel SVM & 15.55 ($\pm$ 6.30) & 53.67 ($\pm$ 6.52) & \textbf{15.13} ($\pm$ 1.44) & \textbf{15.38} ($\pm$ 1.69) \\ 
         
         KNN & 12.78 ($\pm$ 7.80) & 51.00 ($\pm$ 4.86) & 18.56 ($\pm$ 1.36) & 18.30 ($\pm$ 1.82) \\ 
         
         CompRF unsupervised & 11.67 ($\pm$ 7.15) & 38.50 ($\pm$ 4.19) & 17.91 ($\pm$ 1.42) & 19.56 ($\pm$ 1.61)\\ 
         
         CompRF supervised & \textbf{11.11} ($\pm$ 8.28) & \textbf{38.17} ($\pm$ 5.35) & 18.05 ($\pm$ 1.63)  & 18.40 ($\pm$ 2.27) \\ \bottomrule
    \end{tabular}
    \vskip -0.10in
\end{table*}

The results show that, surprisingly, CompRF can slightly outperform the CART forests for classification tasks even though it uses considerably less information. The reason might be that the CompRF splits are better adapted to the geometry of the data than the CART splits. 
While the CART criterion for selecting the exact splitting point can be very  informative for regression (see below), for classification it seems that a simple data dependent splitting criterion as in the supervised CompRF can be as efficient. Conversely, we see that unsupervised splitting as in the unsupervised CompRF is clearly worse than supervised splitting. 

\subsubsection{Regression} 

Next we consider regression tasks on four datasets. Online news popularity (ONP) is a dataset of articles with the popularity of the article as target \citep{fernandes2015proactive}. Boston is a dataset of properties with the estimated value as target variable. ForestFire is a dataset meant to predict the burned area of forest fires, in the northeast region of Portugal \citep{cortez2007data}. WhiteWine (Wine) is a subset of wine quality dataset \citep{cortez2009modeling}. Details of the datasets are shown in the first two rows of Table~\ref{tab:reg_results}.

Since the regression datasets have no separate training and test set, we assign 90\% of the items to the training and the remaining 10\% to the test set. In order to remove the effect of the fixed partitioning, we repeat the experiments 10 times with random training/test set assignments. Note that we use CompRF with unsupervised tree construction for regression.

\textbf{Parameters of CompRF:}  We report the behaviour of the CompRF with respect to the parameters $n_0$ and $M$. We perform 10-fold cross-validation with the same range of parameters as in the previous section. Figure~\ref{fig:RegExp1} shows the average root mean squared error (RMSE) over the 10 folds. The cross-validation is performed for 10 random training/test set assignments. The figure corresponds to the first assignment out of 10 (the behaviour for the  other training/test set assignments is similar). 
The CompRF algorithm shows the best performance with $n_0=1$ for the Boston and ForestFire datasets, however larger values of $n_0$ lead to better performance for other datasets. We believe that the main reason for this variance is the unsupervised tree construction in the CompRF algorithm for regression. 

\textbf{Comparison between CompRF and CART:} Table~\ref{tab:reg_results} shows the average and standard deviation of the RMSE for the CompRF and CART forests over the 10 runs with random training/test set assignment. For each combination of training and test sets we tuned parameters independently by cross validation. CompRF is constructed with unsupervised splitting, while the CART forests are built using a supervised criterion. We can see that on the Boston and Wine datasets, the performance of the CART forest is substantially better than the CompRF. 
In this case, ignoring the Euclidean representation of the data and just relying on the comparison-based trees leads to a significant decrease in performance. However the performance of our method on the other two datasets is quite the same as CART forests. We can conclude that in some cases the CART criterion can be essential for regression. However, note that if we are just given a comparison-based setting---without actual vector space representation---it is hardly possible to propose an efficient supervised criterion for splitting.

\begin{figure*}[ht]
    \vskip 0.2in
    \centering
    \begin{subfigure}
    \centering
    \includegraphics[width=.23\linewidth]{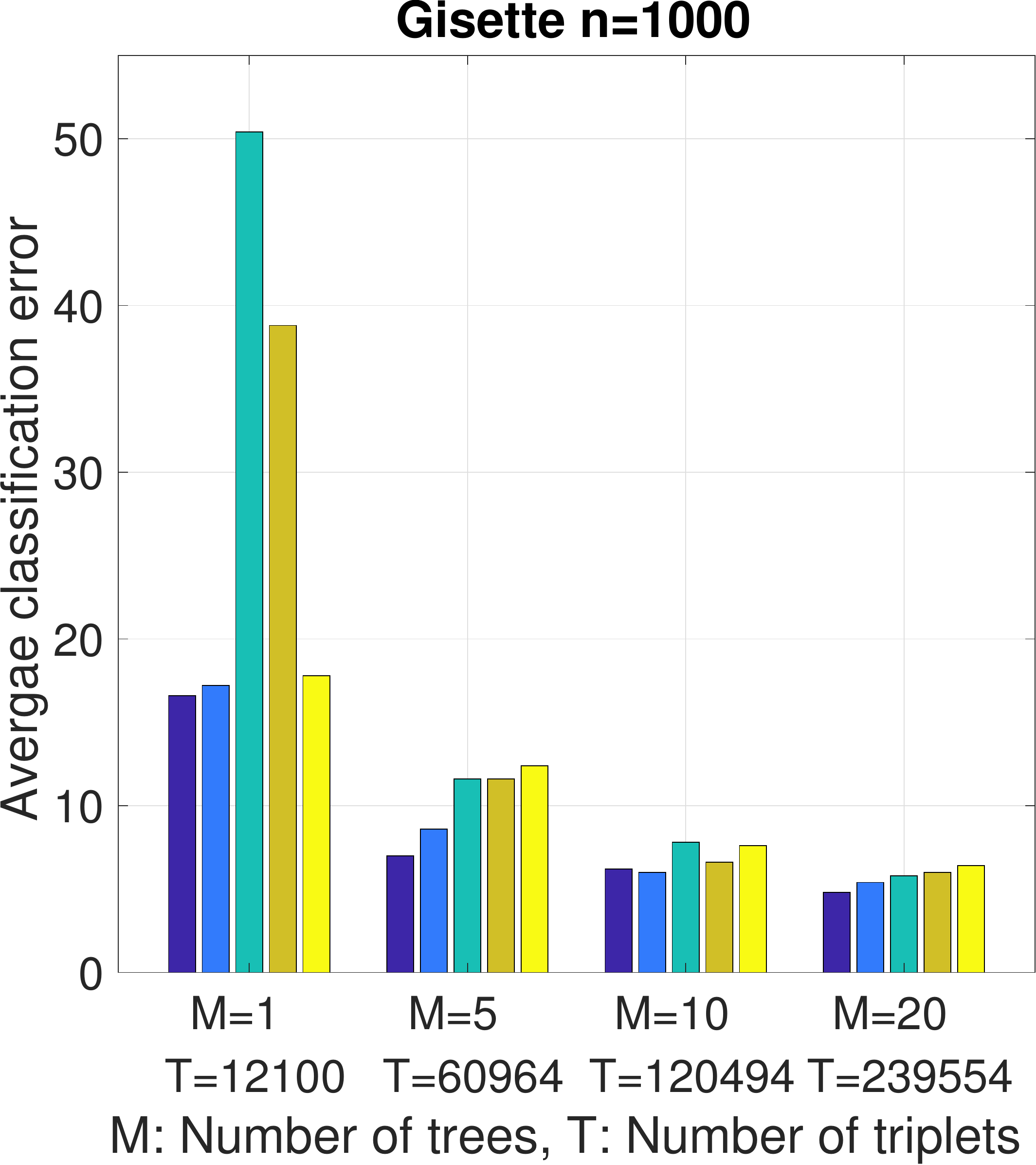}
    \end{subfigure}
    \begin{subfigure}
    \centering
    \includegraphics[width=.23\linewidth]{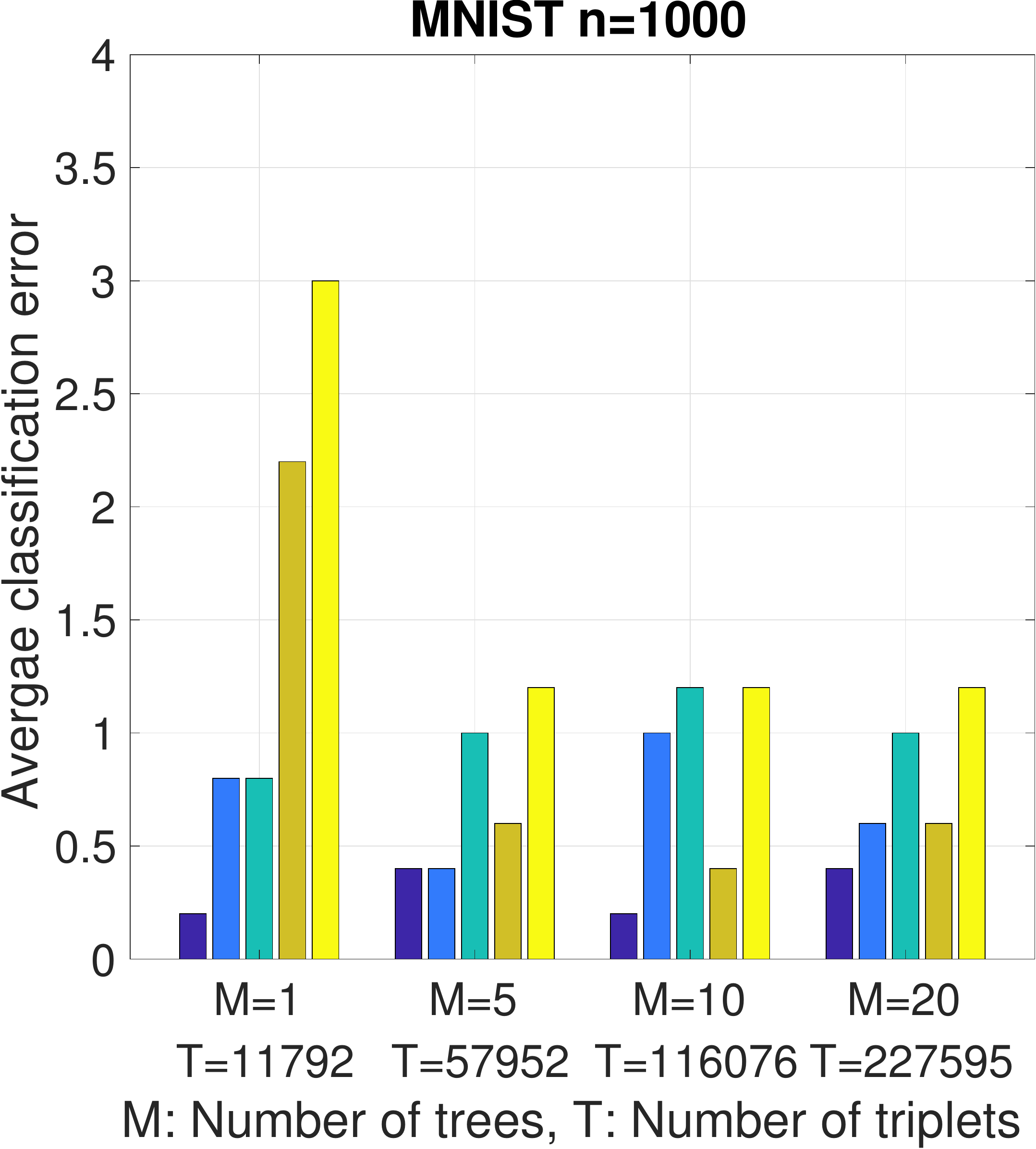}
    \end{subfigure}
    \begin{subfigure}
    \centering
    \includegraphics[width=.23\linewidth]{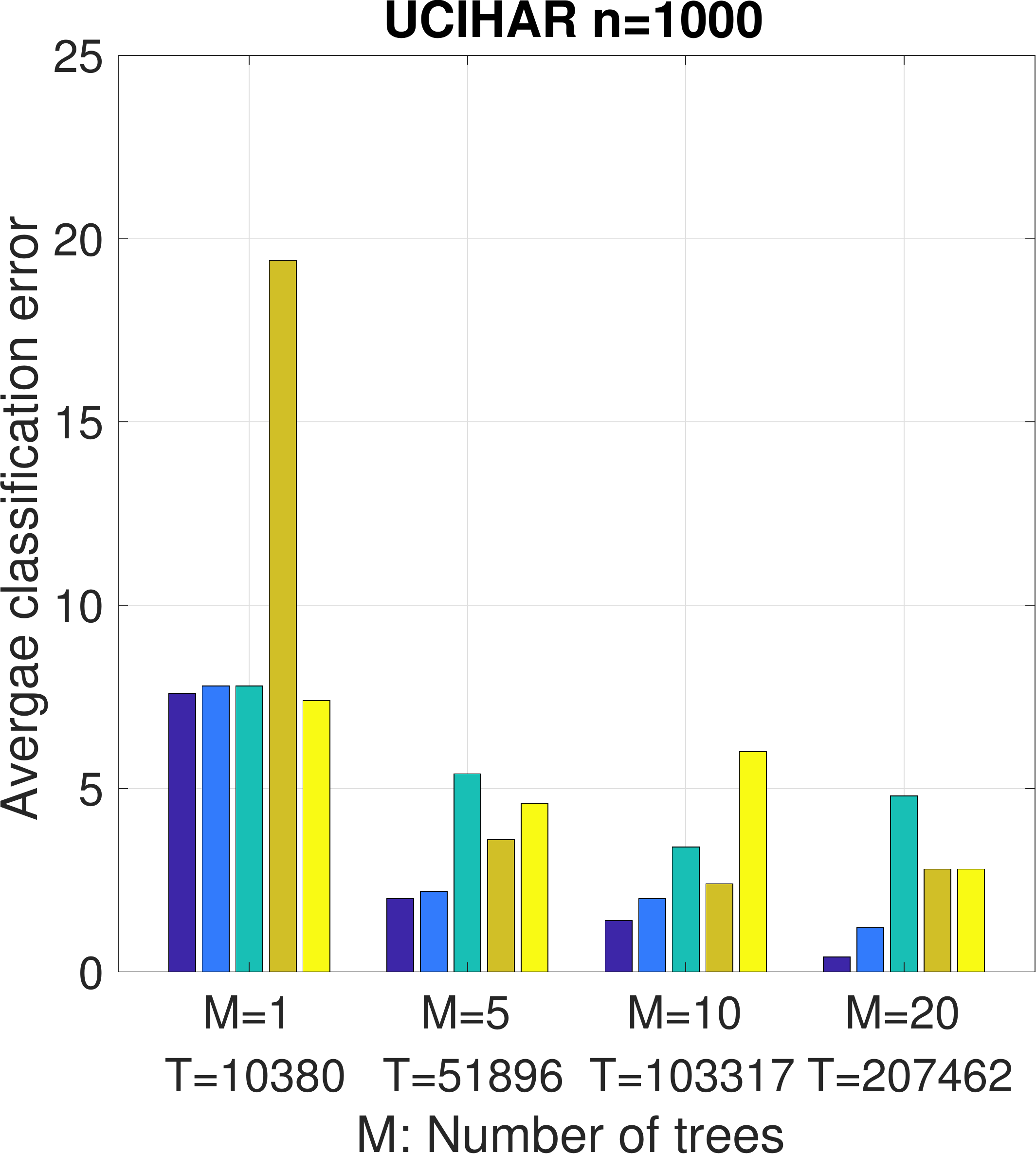}
    \end{subfigure}
    \begin{subfigure}
    \centering
    \includegraphics[width=.225\linewidth]{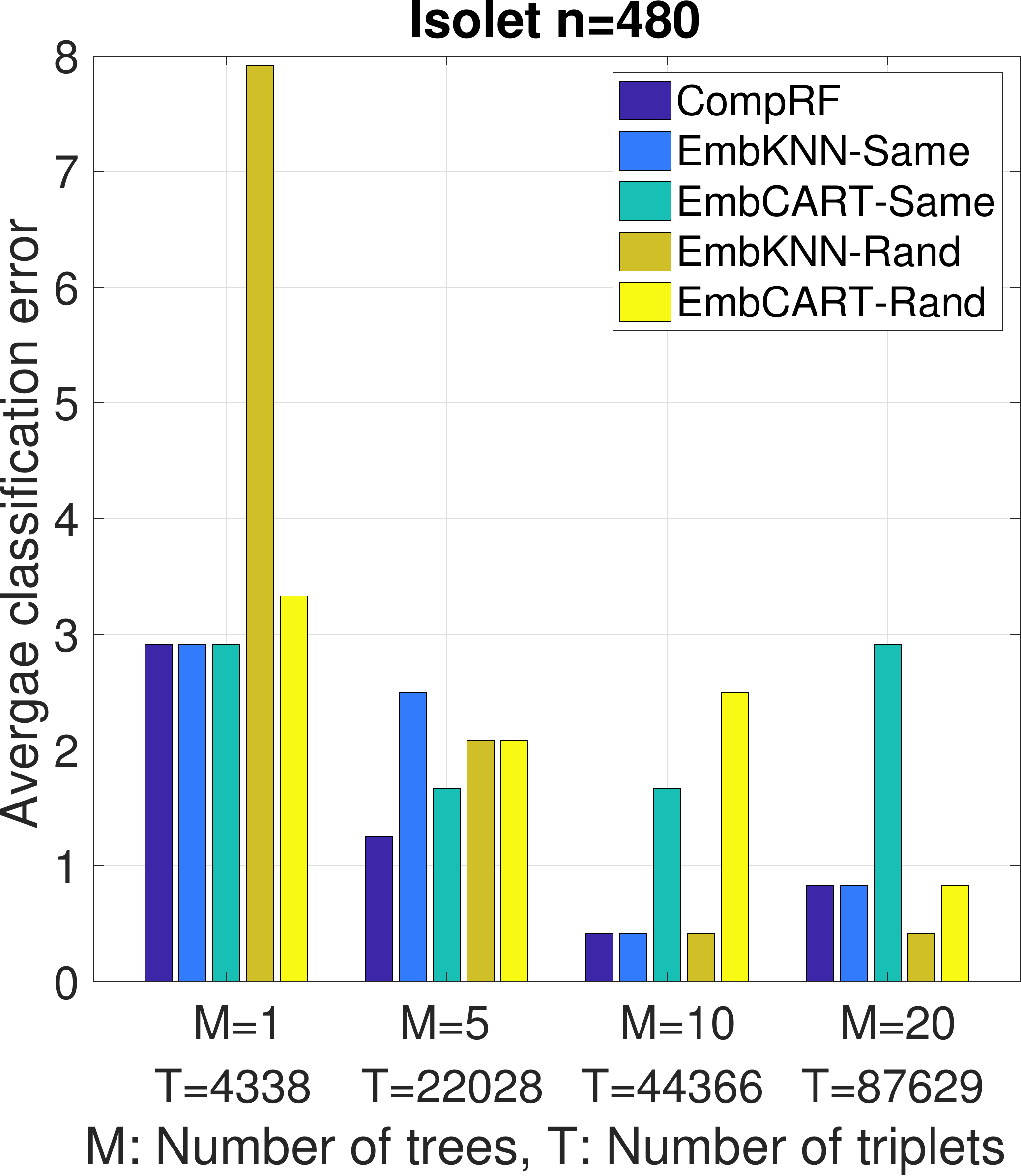}
    \end{subfigure}
    \caption{\label{fig:Set4Exp1}Average classification error of the CompRF in comparison with embedding approach on classification datasets with less than 1000 items. \textit{EmbKNN-Same} (resp. \textit{EmbCART-Same}) denotes the TSTE+KNN using the same triplets as CompRF, while \textit{EmbKNN-Rand} (resp. \textit{EmbCART-Rand}) stands for using TSTE with the same number of random triplets. X-Axis show the number of trees $(M)$ used for the CompRF and the corresponding number of triplets $(T)$ for the embedding. Each set of bars corresponds to a fixed $M$. Note that by increasing $M$, the number of triplets used by CompRF will be increased, as it appears in the X-Axis.}
    \vskip -0.2in
\end{figure*}

\subsection{Metric, non-Euclidean Setting}

In this set of experiments we aim to demonstrate the performance of the CompRF in general metric spaces. We choose graph-structured data for this experiment. Each data-point is a graph, and as a distance between graphs we use graph-based kernel functions. 
In particular, the Weisfeiler-Lehman graph kernels are a family of graph kernels that have promising results on various graph datasets~\citep{shervashidze2011weisfeiler}. We compute the WL-subtree and WL-edge kernels on four of the datasets reported in \citet{shervashidze2011weisfeiler}: MUTAG, ENZYMES, NCI1 and NCI109. In order to evaluate triplet comparisons based on the graph kernels, we first convert the kernel matrix to a distance matrix in the standard way (expressing the Gram matrix in terms of distances). 
 
We compare supervised and unsupervised CompRF with the Kernel SVM and KNN classifier in Table~\ref{tab:kernelbased}. Note that in this setting, CART forests are not applicable as they would require an explicit vector space representation. Parameters of the Kernel SVM and $k$ of the KNN classifier are adjusted with 10-fold cross-validation on training sets.

We set the parameters of the CompRF to $n_0=1$ and $M=200$, as it shows acceptable performance in the Euclidean setting. 
We assign 90\% of the items as training and the remaining 10\% as the test set. The experiment is repeated 10 times with random training/test assignments. The average and standard deviation of classification error is reported in Table~\ref{tab:kernelbased}. The CompRF algorithm outperforms the kernel SVM on the MUTAG and ENZYMES datasets. However, it has slightly lower performance on the other two datasets. However, note that the kernel SVM requires a lot of background knowledge (one has to construct a kernel in the first place, which can be difficult), whereas our CompRF algorithm neither uses the explicit distance values nor requires them to satisfy the axioms of a kernel function. 

\subsection{Comparison-Based Setting}
\label{sec:compbasedExps}

Now we assume that the distance metric is unknown and inaccessible directly, but we can actively ask for triplet comparisons. In this setting, the major competitors to comparison-based forests are indirect methods that first use ordinal embedding to a Euclidean space, and then classify the data in the Euclidean space. As practical active ordinal embedding methods do not really exist we settle for a batch setting in this case. After embedding, we use CART forests and the KNN classifier in the Euclidean space. 

Comparing various ordinal embedding algorithms, such as GNMDS \citep{agarwal2007generalized}, LOE \citep{terada2014local} and TSTE \citep{van2012stochastic} shows that the TSTE in combination with a classifier consistently outperforms the others (see Section~3.1 in the supplement). Therefore, we here only report the comparison with the TSTE embedding algorithm.
We choose the embedding dimension by 2-fold cross-validation in the range of $d \in \{10,20,30,40,50\}$ (embedding in more than 50 dimensions is impossible in practice due to the running time of the TSTE). We also adjust $k$ of the KNN classifier in the cross-validation process.

We design a comparison-based scenario based on Euclidean datasets. First, we let CompRF choose the desired triplets to construct the forest and classify the test points. The embedding methods are used in two different scenarios: once with exactly the same triplets as in the CompRF algorithm, and once with a completely random set of triplets of the same size as the one used by CompRF. 

The size of our datasets by far exceeds the number of points that embedding algorithms, particularly TSTE, can handle. To reduce the size of the datasets, we choose the first two classes, then we subsample $1000$ items. Isolet has already less than $1000$ items in first two classes.
We assign half of the dataset as training and the other half as test set. Bar plots in Figure~\ref{fig:Set4Exp1} show the classification error of the CompRF in comparison with embedding methods with various numbers of trees in the forests ($M$). 
We set $n_0=1$ for the CompRF. 

In each set of bars, which corresponds to a restricted comparison-based regime, CompRF outperforms embedding methods or has the same performance. Another significant advantage of CompRF in comparison with the embedding is the low computation cost. A simple demonstration is provided in Section~3.3 of the supplementary material.

\section{Conclusion and Future Work}

We propose comparison-based forests for classification and regression tasks. This method only requires comparisons of distances as input. 
From a practical point of view, it works surprisingly well in all kinds of circumstances (Euclidean spaces, metric spaces, comparison-based setting) and is much simpler and more efficient than some of its competitors such as ordinal embeddings.

We have proven consistency in a simplified setting. As future work, this analysis should be extended to more realistic situations, namely tree construction depending on the sample; 
forests with inconsistent trees, but the forest is still consistent; and finally the supervised splits. In addition, it would be interesting to propose a comparison-based supervised tree construction for the regression tasks. 

\clearpage

\section*{Acknowledgements}

The authors thank Debarghya Goshdastidar and Micha\"el Perrot for fruitful discussions. This work has been supported by the German Research Foundation DFG (SFB 936/ Z3), the Institutional Strategy of the University of T{\"u}bingen (DFG ZUK 63), and the International Max Planck Research School for
Intelligent Systems (IMPRS-IS). 

\bibliographystyle{plainnat} 
\bibliography{refs}

\begin{thebibliography}{34}
\providecommand{\natexlab}[1]{#1}
\providecommand{\url}[1]{\texttt{#1}}
\expandafter\ifx\csname urlstyle\endcsname\relax
  \providecommand{\doi}[1]{doi: #1}\else
  \providecommand{\doi}{doi: \begingroup \urlstyle{rm}\Url}\fi

\bibitem[Agarwal et~al.(2007)Agarwal, Wills, Cayton, Lanckriet, Kriegman, and
  Belongie]{agarwal2007generalized}
S.~Agarwal, J.~Wills, L.~Cayton, G.~Lanckriet, D.~Kriegman, and S.~Belongie.
\newblock Generalized non-metric multidimensional scaling.
\newblock In \emph{AISTATS}, pages 11--18, 2007.

\bibitem[Balcan et~al.(2016)Balcan, Vitercik, and White]{balcan2016Learning}
M.F. Balcan, E.~Vitercik, and C.~White.
\newblock Learning combinatorial functions from pairwise comparisons.
\newblock In \emph{COLT}, pages 310--335, 2016.

\bibitem[Biau(2012)]{Bia:2012}
G.~Biau.
\newblock Analysis of a random forests model.
\newblock \emph{JMLR}, 13\penalty0 (4):\penalty0 1063--1095, 2012.

\bibitem[Biau and Scornet(2016)]{biau2016random}
G.~Biau and E.~Scornet.
\newblock A random forest guided tour.
\newblock \emph{Test}, 25\penalty0 (2):\penalty0 197--227, 2016.

\bibitem[Biau et~al.(2008)Biau, Devroye, and Lugosi]{Bia_Dev_Lug:2008}
G.~Biau, L.~Devroye, and G.~Lugosi.
\newblock Consistency of random forests and other averaging classifiers.
\newblock \emph{JMLR}, 9\penalty0 (9):\penalty0 2015--2033, 2008.

\bibitem[Breiman(1996)]{breiman1996bagging}
L.~Breiman.
\newblock Bagging predictors.
\newblock \emph{Machine learning}, 24\penalty0 (2):\penalty0 123--140, 1996.

\bibitem[Breiman(2001)]{breiman2001random}
L.~Breiman.
\newblock Random forests.
\newblock \emph{Machine Learning}, 45\penalty0 (1):\penalty0 5--32, 2001.

\bibitem[Breiman et~al.(1984)Breiman, Friedman, Stone, and
  Olshen]{breiman1984classification}
L.~Breiman, J.~Friedman, C.J. Stone, and R.A. Olshen.
\newblock \emph{Classification and regression trees}.
\newblock CRC press, 1984.

\bibitem[Cortez and Morais(2007)]{cortez2007data}
P.~Cortez and A.J.R. Morais.
\newblock {A Data Mining Approach to Predict Forest Fires using Meteorological
  Data}.
\newblock In \emph{Portuguese Conference on Artificial Intelligence}, pages
  512--523, 2007.

\bibitem[Cortez et~al.(2009)Cortez, Cerdeira, Almeida, Matos, and
  Reis]{cortez2009modeling}
P.~Cortez, A.~Cerdeira, F.~Almeida, T.~Matos, and J.~Reis.
\newblock Modeling wine preferences by data mining from physicochemical
  properties.
\newblock \emph{Decision Support Systems}, 47\penalty0 (4):\penalty0 547--553,
  2009.

\bibitem[Dasgupta and Freund(2008)]{dasgupta2008random}
S.~Dasgupta and Y.~Freund.
\newblock Random projection trees and low dimensional manifolds.
\newblock In \emph{STOC}, pages 537--546, 2008.

\bibitem[Denil et~al.(2013)Denil, Matheson, and Freitas]{Den_Mat_Fre:2013}
M.~Denil, D.~Matheson, and N.~Freitas.
\newblock Consistency of online random forests.
\newblock In \emph{ICML}, pages 1256--1264, 2013.

\bibitem[Devroye et~al.(1996)Devroye, Gy{\"o}rfi, and Lugosi]{Dev_Gyo_Lug:1996}
L.~Devroye, L.~Gy{\"o}rfi, and G.~Lugosi.
\newblock \emph{A probabilistic theory of pattern recognition}.
\newblock Springer, 1996.

\bibitem[Ezra and Sharir(2017)]{ezra2017nearly}
E.~Ezra and M.~Sharir.
\newblock A nearly quadratic bound for the decision tree complexity of k-sum.
\newblock In \emph{LIPIcs-Leibniz International Proceedings in Informatics},
  volume~77. Schloss Dagstuhl-Leibniz-Zentrum fuer Informatik, 2017.

\bibitem[Fernandes et~al.(2015)Fernandes, Vinagre, and
  Cortez]{fernandes2015proactive}
K.~Fernandes, P.~Vinagre, and P.~Cortez.
\newblock A proactive intelligent decision support system for predicting the
  popularity of online news.
\newblock In \emph{Portuguese Conference on Artificial Intelligence}, pages
  535--546, 2015.

\bibitem[Fern{\'a}ndez-Delgado et~al.(2014)Fern{\'a}ndez-Delgado, Cernadas,
  Barro, and Amorim]{fernandez2014we}
M.~Fern{\'a}ndez-Delgado, E.~Cernadas, S.~Barro, and D.~Amorim.
\newblock Do we need hundreds of classifiers to solve real world classification
  problems.
\newblock \emph{JMLR}, 15\penalty0 (1):\penalty0 3133--3181, 2014.

\bibitem[Haghiri et~al.(2017)Haghiri, Ghoshdastidar, and von
  Luxburg]{haghiri2017comparison}
S.~Haghiri, D.~Ghoshdastidar, and U.~von Luxburg.
\newblock Comparison-based nearest neighbor search.
\newblock In \emph{AISTATS}, pages 851--859, 2017.

\bibitem[Heikinheimo and Ukkonen(2013)]{heikinheimo2013crowd}
H.~Heikinheimo and A.~Ukkonen.
\newblock The crowd-median algorithm.
\newblock In \emph{HCOMP}, 2013.

\bibitem[Ishwaran and Kogalur(2010)]{Ish_Kog:2010}
H.~Ishwaran and U.~B. Kogalur.
\newblock Consistency of random survival forests.
\newblock \emph{Statistics \& probability letters}, 80\penalty0 (13):\penalty0
  1056--1064, 2010.

\bibitem[Kane et~al.(2017{\natexlab{a}})Kane, Lovett, and Moran]{kane2017near}
D.M. Kane, S.~Lovett, and S.~Moran.
\newblock Near-optimal linear decision trees for k-sum and related problems.
\newblock \emph{arXiv preprint arXiv:1705.01720}, 2017{\natexlab{a}}.

\bibitem[Kane et~al.(2017{\natexlab{b}})Kane, Lovett, Moran, and
  Zhang]{kane2017active}
D.M. Kane, S.~Lovett, S.~Moran, and J.~Zhang.
\newblock Active classification with comparison queries.
\newblock \emph{Foundations Of Computer Science (FOCS)}, 2017{\natexlab{b}}.

\bibitem[Kleindessner and Luxburg(2015)]{kleindessner2015dimensionality}
M.~Kleindessner and U.~Luxburg.
\newblock Dimensionality estimation without distances.
\newblock In \emph{AISTATS}, pages 471--479, 2015.

\bibitem[Kleindessner and von Luxburg(2017)]{kleindessner2017kernel}
M.~Kleindessner and U.~von Luxburg.
\newblock Kernel functions based on triplet comparisons.
\newblock In \emph{NIPS}, pages 6810--6820, 2017.

\bibitem[LeCun et~al.(1998)LeCun, Bottou, Bengio, and Haffner]{mnist}
Y.~LeCun, L.~Bottou, Y.~Bengio, and P.~Haffner.
\newblock Gradient-based learning applied to document recognition.
\newblock \emph{Proceedings of the IEEE}, 86\penalty0 (11):\penalty0
  2278--2324, 1998.

\bibitem[Lichman(2013)]{Lichman:2013}
M.~Lichman.
\newblock {UCI} machine learning repository, 2013.
\newblock Available at \url{http://archive.ics.uci.edu/ml}.

\bibitem[Scornet et~al.(2015)Scornet, Biau, and Vert]{Sco_Bia_Ver:2015}
E.~Scornet, G.~Biau, and J.-P. Vert.
\newblock Consistency of random forests.
\newblock \emph{The Annals of Statistics}, 43\penalty0 (4):\penalty0
  1716--1741, 2015.

\bibitem[Shervashidze et~al.(2011)Shervashidze, Schweitzer, Leeuwen, Mehlhorn,
  and Borgwardt]{shervashidze2011weisfeiler}
N.~Shervashidze, P.~Schweitzer, E.J. Leeuwen, K.~Mehlhorn, and K.M. Borgwardt.
\newblock Weisfeiler-lehman graph kernels.
\newblock \emph{JMLR}, 12:\penalty0 2539--2561, 2011.

\bibitem[Tamuz et~al.(2011)Tamuz, Liu, Belongie, Shamir, and
  Kalai]{tamuz2011adaptively}
O.~Tamuz, C.~Liu, S.~Belongie, O.~Shamir, and A.~Kalai.
\newblock Adaptively learning the crowd kernel.
\newblock In \emph{ICML}, pages 673--680, 2011.

\bibitem[Terada and von Luxburg(2014)]{terada2014local}
Y.~Terada and U.~von Luxburg.
\newblock Local ordinal embedding.
\newblock In \emph{ICML}, pages 847--855, 2014.

\bibitem[Ukkonen et~al.(2015)Ukkonen, Derakhshan, and
  Heikinheimo]{ukkonen2015crowdsourced}
A.~Ukkonen, B.~Derakhshan, and H.~Heikinheimo.
\newblock Crowdsourced nonparametric density estimation using relative
  distances.
\newblock In \emph{HCOMP}, 2015.

\bibitem[van~der Maaten and Weinberger(2012)]{van2012stochastic}
L.~van~der Maaten and K.~Weinberger.
\newblock Stochastic triplet embedding.
\newblock In \emph{Machine Learning for Signal Processing (MLSP)}, pages 1--6,
  2012.
\newblock Code available at \url{http://homepage.tudelft.nl/19j49/ste}.

\bibitem[Wah et~al.(2015)Wah, Maji, and Belongie]{wah2015learning}
C.~Wah, S.~Maji, and S.~Belongie.
\newblock Learning localized perceptual similarity metrics for interactive
  categorization.
\newblock In \emph{Winter Conference on Applications of Computer Vision
  (WACV)}, pages 502--509, 2015.

\bibitem[Wilber et~al.(2015)Wilber, Kwak, Kriegman, and
  Belongie]{wilber2015learning}
M.~Wilber, I.S. Kwak, D.~Kriegman, and S.~Belongie.
\newblock Learning concept embeddings with combined human-machine expertise.
\newblock In \emph{ICCV}, pages 981--989, 2015.

\bibitem[Zhang et~al.(2015)Zhang, Maji, and Tomioka]{zhang2015jointly}
L.~Zhang, S.~Maji, and R.~Tomioka.
\newblock Jointly learning multiple measures of similarities from triplet
  comparisons.
\newblock \emph{arXiv preprint arXiv:1503.01521}, 2015.

\end{thebibliography}

\clearpage



\section*{Supplementary material}

<div style="text-align:center">

Supplementary material for the article:

# Comparison-Based Random Forests

</div>

<div style="text-align:center">

Siavash Haghiri, Damien Garreau, Ulrike von Luxburg

</div>

In this supplementary material, we provide a complete proof of Theorem 2 and additional experiments. The main arguments of this proof are collected in Section 1, while all auxiliary results can be found in Section 2. Section 3 contains the supplemental experiments.

## 1  Main proofs

Let us recall our main assumptions on the data and the central result regarding the consistency of the continuous comparison tree.

**Assumption 1.1** (**Bounded density on the unit cube**)**.** The random variable $X \in [0,1]^d$ has density $f$ with respect to the Lebesgue measure on $[0,1]^d$. Additionally, there exist constants $0 < f_{\min} \le f_{\max} < +\infty$ such that

$$\forall x \in \mathcal{X}, \quad f_{\min} \le f(x) \le f_{\max}.$$

**Theorem 1.1** (**Consistency of comparison trees**)**.** *Assume that $X$ satisfies Assumption 1.1. Then the partitioning classification rule associated to $T^0(\alpha \log n)$ is consistent for any $0 < \alpha < 1/\log 2$.*

As explained in the paper, the proof of Theorem 1.1 follows from Devroye et al. (1996, Theorem 6.1). It first requires control on the number of samples in the leaves of the tree. This is resolved in Section 1.2. Second, we need to bound the diameter of the leaves of the tree, which is achieved in Section 1.3. Before turning to these proofs, we state and prove the key result used to control the diameter (Proposition 1.1) in the next section.

Let us now introduce some additional notation. For any convex compact subset $A$, we define $\pi_A$ as the orthogonal projection on $A$. For any given $x \in \mathbb{R}^d$ and $r > 0$, we define $\mathcal{B}(x, r)$ as the closed ball of center $x$ and radius $x$. Namely,

$$\mathcal{B}(x, r) := \left\{ y \in \mathbb{R}^d \mid \|x - y\| \le r \right\}.$$

The sphere of center $x$ and radius $r$ is the boundary of $\mathcal{B}(x, r)$, that is,

$$\mathcal{S}(x, r) := \left\{ y \in \mathbb{R}^d \mid \|x - y\| = r \right\}.$$

When working in dimension 2, $\mathcal{S}(x, r)$ is simply the circle of center $x$ and radius $r$, denoted by $\mathcal{C}(x, r)$. We call *annulus* the (closed) set of points comprised between two concentric spheres, that is, for any $x \in \mathbb{R}^d$, $r_1, r_2 > 0$,

$$\mathcal{A}(x, r_1, r_2) := \left\{ y \in \mathbb{R}^d \mid r_1 \le \|x - y\| \le r_2 \right\}.$$

If $r_1 \le 0$, we set $\mathcal{A}(x, r_1, r_2) = \mathcal{B}(a, r_2)$.

## 1.1 Control of the cell diameter

Let us recall Proposition 1.1, the key result needed for proving that the diameter of the comparison-tree leaves goes to zero in probability.

**Proposition 1.1** (**Diameter control**). *Assume that Assumption 1.1 holds. Let $C$ be a cell of $T^0(X)$ such that $\mathrm{diam}\,(C) \leq D$. Then the probability that there exists a descendant of $C$ which is more than $k$ levels below and yet has diameter greater than $D/2$ is at most $N_{f,d}(N_{f,d} + 1)\gamma_{f,d}^k/2$, where $0 < N_{f,d}$ and $0 < \gamma_{f,d} < 1$ are constants depending only on $d$, $f_{\min}$, and $f_{\max}$.*

Proposition 1.1 is an analogous of Lemma 12 in Dasgupta and Freund (2008). In plain words, it states that for any cell of the continuous comparison tree, the diameter of any descendant at least $k$ levels below is halved with high probability depending on $k$. Our proof follows closely that of Dasgupta and Freund (2008, Lemma 12), the main difference being in the auxiliary lemmas used to control the probability of certain events, due to the radically different nature of the random tree that we consider.

*Proof.* Consider a cover of $C$ by balls of radius $r = D/c_r$, with $c_r := 2^6 \cdot d \cdot 25^d \cdot \frac{f_{\max}^2}{f_{\min}^2}$. According to Shalev-Shwartz and Ben-David (2014, Section 27.1), at most

$$\left(\frac{2D\sqrt{d}}{r}\right)^d = \left(2^7 \cdot d^{3/2} \cdot 25^d \cdot \frac{f_{\max}^2}{f_{\min}^2}\right)^d =: N_{d,f}$$

such balls are needed, since $\mathrm{diam}\,(C) \leq D$.

Fix any pair of balls $B, B'$ from this cover whose centers are at distance at least $D/2 - r$ from one another. Given any $x$ and $y$, we say that the split according to $\Delta(x, y)$ is a *good cut* if it cleanly separates $B$ from $B'$, i.e., if $B \subset H_x$ and $B' \subset H_y$ or $B' \subset H_x$ and $B \subset H_y$. If the split cuts both $B$ and $B'$, that is, $B \cap \Delta(x, y) \neq \emptyset$ and $B' \cap \Delta(x, y) \neq \emptyset$, we say that it is a *bad cut*. See Figure 1 for illustration.

For any $k \geq 1$, let $p_k$ be the probability that there is some cell $k$ levels below $C$ which contains points from both $B$ and $B'$. We write

$$\begin{aligned}
p_k &\leq \mathbb{P}\,(\text{top split is a good cut}) \cdot 0 + \mathbb{P}\,(\text{top split is a bad cut}) \cdot 2p_{k-1} \\
&\quad + \mathbb{P}\,(\text{all other split configurations}) \cdot p_{k-1} \\
&\leq (1 + \mathbb{P}\,(\text{top split is a bad cut}) - \mathbb{P}\,(\text{top split is a good cut}))\, p_{k-1}.
\end{aligned}$$

Since $d \geq 1$ and $c_r > 50$, according to Lemma 2.1 and 2.2,

$$\begin{aligned}
\mathbb{P}\,(\text{top split is a bad cut}) - \mathbb{P}\,(\text{top split is a good cut}) &\leq \frac{f_{\max}}{f_{\min}} \cdot \frac{64d}{c_r} - 2 \cdot \frac{f_{\min}}{f_{\max}} \cdot \frac{1}{25^d} \\
&= -\frac{f_{\min}}{f_{\max}} \cdot \frac{1}{25^d} < 0.
\end{aligned}$$

Set $\gamma_{f,d} := 1 - \frac{f_{\min}}{f_{\max}} \cdot \frac{1}{25^d}$, we just showed that

$$p_k \leq \gamma_{f,d} p_{k-1}. \tag{1.1}$$

Since $p_0 = 1$, we deduce that $p_k \leq \gamma_{f,d}^k$. We conclude by a union bound over all the pairs from the cover that are at the prescribed minimum distance from each other. $\qquad\square$

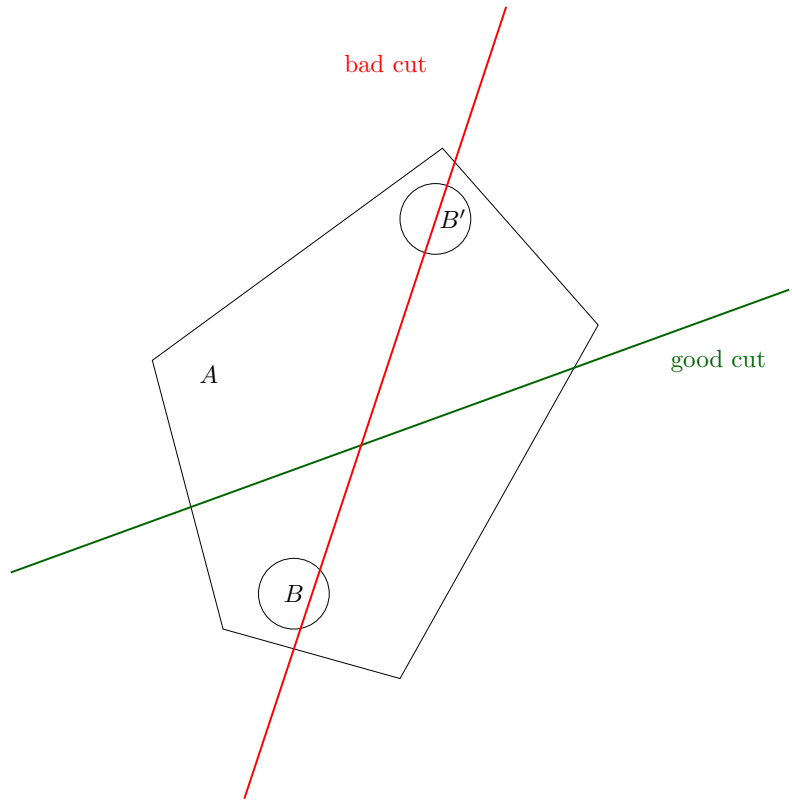

Figure 1: Good cuts and bad cuts. The current cell $A$ contains $B$ and $B'$, two faraway balls of small radius—with respect to the diameter of $A$. A *good cut* (in green) cleanly separates $B$ and $B'$, whereas a *bad cut* (in red) intersects both.

Note that the main difference with the proof of Lemma 12 in Dasgupta and Freund (2008) comes from Eq. (1.1). Namely, in the setting of Dasgupta and Freund (2008), $\gamma_{f,d}$ is a constant that does not depend on the dimension. The dependency on the dimension in our case is due to the lower bound on the probability of a good cut that we obtain in Lemma 2.1, which is decreasing exponentially with the dimension. Improving this bound, namely obtaining a bound without exponential dependency in the dimension, would yield a "more reasonable" number of levels required to divide the diameter by two in Proposition 1.1.

## 1.2   $N(X) \longrightarrow +\infty$ in probability

According to Lemma 20.1 in Devroye et al. (1996) and the remark that follows, it is sufficient to show that the number of regions is o $(n)$. For each $n$, by construction, $T^0(\alpha \log n)$ has $2^{\alpha \log n} = n^{\alpha \log 2}$ leafs. Since $\alpha \log 2 < 1$, $2^{\alpha \log n} = \mathrm{o}\,(n)$ as $n \to +\infty$. $\qquad \square$

## 1.3   diam $(A(X)) \longrightarrow 0$ in probability

Let $0 < \varepsilon < 1$. In this section, we show that

$$\mathbb{P}\left(\operatorname{diam}\left(A(X)\right) > \varepsilon\right) \longrightarrow 0 \quad \text{when} \quad n \to +\infty \,.$$

We first notice that

$$\mathbb{P}\left(\operatorname{diam}\left(A(X)\right) > \varepsilon\right) \le \max_i \mathbb{P}\left(\operatorname{diam}\left(A_{i,n}\right) > \varepsilon\right) \,.$$

Let $A$ be the leaf of $T^0(p_n)$ with maximal diameter and define $\pi := \left\lceil \frac{\log(\sqrt{d}) - \log \varepsilon}{\log 2} \right\rceil$, so that $\varepsilon > \sqrt{d}/2^\pi$. We write

$$\mathbb{P}\left(\operatorname{diam}\left(A\right) > \varepsilon\right) \le \mathbb{P}\left(\operatorname{diam}\left(A\right) > \frac{\sqrt{d}}{2^\pi}\right) \,.$$

Define $C_1, \ldots, C_{p_n}$ the path from $C_0 = [0,1]^d$ to $C_{p_n} = A$ in the tree $T^0(p_n)$. Set $k = \left\lfloor \frac{p_n}{\pi} \right\rfloor$. Set $A^{(0)} = C_0$, $A^{(1)} = C_k$, $A^{(2)} = C_{2k}$, …, $A^{(\pi-1)} = C_{(\pi-1)k}$ and $A^{(\pi)} = A$. We define the event $E_j := \left\{\operatorname{diam}\left(A^{(j)}\right) > \sqrt{d}/2^j\right\}$. Then

$$\mathbb{P}\left(\operatorname{diam}\left(A\right) > \frac{\sqrt{d}}{2^\pi}\right) = \mathbb{P}\left(E_\pi \middle| E_{\pi-1}\right) \cdot \mathbb{P}\left(E_{\pi-1}\right) + \mathbb{P}\left(E_\pi \middle| E_{\pi-1}^{\mathsf{c}}\right) \cdot \mathbb{P}\left(E_{\pi-1}^{\mathsf{c}}\right)$$

$$\text{(law of total probability)}$$

$$\le \mathbb{P}\left(E_\pi \middle| E_{\pi-1}^{\mathsf{c}}\right) + \mathbb{P}\left(E_{\pi-1}\right) \,.$$

Repeating $\pi$ times this reasoning, and since diam $\left(A^{(0)}\right) \le \sqrt{d}$ almost surely, we deduce that

$$\mathbb{P}\left(\operatorname{diam}\left(A^{(\pi)}\right) > \varepsilon\right) \le \sum_{t=1}^{\pi} \mathbb{P}\left(\operatorname{diam}\left(A^{(t)}\right) > \frac{\sqrt{d}}{2^t} \middle| \operatorname{diam}\left(A^{(t-1)}\right) \le \frac{\sqrt{d}}{2^{t-1}}\right) \,.$$

There are always more than $k$ levels between $A^{(tk)}$ and $A^{((t-1)k)}$ by construction. Hence, according to Proposition 1.1,

$$\mathbb{P}\left(\operatorname{diam}\left(A\right) > \varepsilon\right) \le \pi \cdot \frac{N_{f,d}(N_{f,d} + 1)}{2} \cdot \gamma_{f,d}^k \,.$$

Since $k = O\left(\log n\right)$ and $\gamma_{f,d} \in (0,1)$, we can conclude the proof. $\qquad \square$

## 2 Auxiliary results

The key in proving Proposition 1.1 is to show that, for a given cell, the probability of a "good cut" is greater than the probability of a "bad cut." We thus proceed to prove a lower bound for the probability of a good cut (Section 2.1) and an upper bound for the probability of a bad cut (Section 2.2). Since the first cell is the unit cube and all subsequent cells are obtained by intersection with half-spaces, note that any cell of the comparison tree is a full-dimensional convex polytope almost surely. Thus we state and prove our results for such objects.

### 2.1 Good cuts

The following Lemma is an analogous of Lemma 10 in Dasgupta and Freund (2008). It provides a lower bound on the probability of cleanly separating faraway balls.

**Lemma 2.1** (**Probability of good cut is lower bounded**). *Suppose that Assumption 1.1 holds. Let $A \subset \mathbb{R}^d$ be a full-dimensional convex polytope such that* $\mathrm{diam}\,(A) \leq D < +\infty$. *Let $c_r > 10$ be a constant. Pick any two balls $B := \mathcal{B}\,(z, r)$ and $B' := \mathcal{B}\,(z', r)$ such that*

*(i) both $B$ and $B'$ intersect $A$;*

*(ii) their radius is at most $D/c_r$;*

*(iii) the distance between their centers satisfies $\|z - z'\| \geq D/2 - r$.*

*Then, if $X_1$ and $X_2$ are chosen independently from $A$ according to the distribution of $X$,*

$$\mathbb{P}\,(A \cap B \subset A \cap H_{X_1} \ and \ A \cap B' \subset A \cap H_{X_2}) \geq 2 \frac{f_{\min}}{f_{\max}} \left( \frac{c_r - 10}{4c_r} \right)^{2d}.$$

*As a direct consequence, if $c_r > 50$,*

$$\mathbb{P}\,(A \cap B \subset A \cap H_{X_1} \ and \ A \cap B' \subset A \cap H_{X_2}) \geq \frac{f_{\min}}{f_{\max}} \frac{2}{25^d}.$$

While the statement of Lemma 2.1 is close to that of of Lemma 10 in Dasgupta and Freund (2008), a major difference lies in the quality of the bound we obtain. Indeed, our bound depends exponentially in the dimension, therefore becoming arbitrarily loose for large values of $d$.

*Proof.* The proof follows the following scheme. First, we conveniently restrict ourselves to the case where the centers of $B$ and $B'$ both belong to $A$ by geometric arguments. We then use Lemma 2.5 to lower bound the probability of a good split by the probability that $x$ and $y$ belong to certain balls $\gamma$ and $\gamma'$. We conclude the proof by finding an upper bound for the volume of $A$ and a lower bound for the volume of $\gamma \cap A$. We refer to Figure 2 throughout this proof.

**Preliminary computations.** Set $a := \pi_A(z)$, $a' := \pi_A(z')$, $\beta := \mathcal{B}\,(a, r)$, and $\beta' := \mathcal{B}\,(a', r)$. Then, according to Lemma 2.3, $A \cap B \subset \beta$ and $A \cap B' \subset \beta'$. For any $x, y \in A$ such that $\beta \subset H_x$ and $\beta' \subset H_y$. Since $A \cap B \subset \beta$, we have $A \cap B \subset H_x$. Furthermore, $A \cap B \subset A$, thus $A \cap B \subset A \cap H_x$. A similar reasoning shows that $A \cap B' \subset A \cap H_y$. Hence

$$\mathbb{P}\,(A \cap B \subset A \cap H_{X_1} \text{ and } A \cap B' \subset A \cap H_{X_2}) \geq \mathbb{P}\,(\beta \subset H_{X_1} \text{ and } \beta' \subset H_{X_2})\,.$$

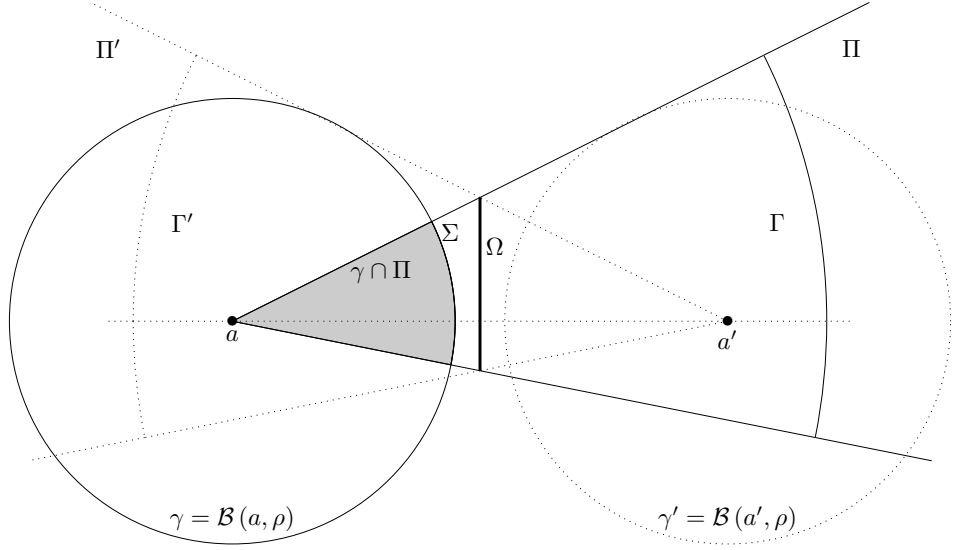

Figure 2: Construction of $\Omega$, $\Pi$ and $\Sigma$. The central thick line represents $\Omega$, the intersection between $A$ and the hyperplane $\Delta(a, a')$. The half-cone $\Pi$ is the union for all $\omega \in \Omega$ of the half-lines $[a, \omega)$. Finally, the spherical cap $\Sigma$ is defined as the intersection between $\mathcal{S}(a, \rho)$ and $\Pi$. In dotted lines we draw the counter-parts of these objects for $a'$. The gray area represents $\gamma \cap \Pi$.

Set $\delta := \|a - a'\|$. Since $a \in B$ and $a' \in B'$, by the triangle inequality, $\|a - a'\| \geq \|z - z'\| - 2r$. By hypothesis, $\|z - z'\| \geq D/2 - r$ and $r \leq D/c_r$, thus

$$\|a - a'\| \geq \frac{D}{2} - 3r \geq \frac{c_r - 6}{2c_r} \cdot D \,.$$

Define $\rho := \|a - a'\| / 2 - r$. We have $\rho \geq \frac{c_r - 10}{4c_r} \cdot D$. In particular, as $c_r > 10$, $\rho > 0$. Then, according to Lemma 2.5,

$$\mathbb{P}\left(\beta \subset H_{X_1} \text{ and } \beta' \subset H_{X_2}\right) \geq \mathbb{P}\left(X_1 \in \gamma \text{ and } X_2 \in \gamma' \text{ or } X_2 \in \gamma \text{ and } X_1 \in \gamma'\right) \,,$$

where $\gamma := \mathcal{B}(a, \rho)$ and $\gamma' := \mathcal{B}(a', \rho)$. Since $X_1$ and $X_2$ are independent and identically distributed and $\gamma \cap \gamma' = \emptyset$,

$$\mathbb{P}\left(X_1 \in \gamma \text{ and } X_2 \in \gamma' \text{ or } X_2 \in \gamma \text{ and } X_1 \in \gamma'\right) \geq 2\,\mathbb{P}\left(X_1 \in \gamma\right)\mathbb{P}\left(X_2 \in \gamma'\right) \,.$$

Since we sample $X_1$ and $X_2$ according to the law of $X$ restricted to $A$ and since Assumption 1.1 holds,

$$\mathbb{P}\left(X_1 \in \gamma\right) \geq \frac{f_{\min}}{f_{\max}} \frac{\mathrm{Vol}_d\left(\gamma \cap A\right)}{\mathrm{Vol}_d\left(A\right)} \,.$$

In the next paragraphs, we find an upper bound for $\mathrm{Vol}_d\left(A\right)$ and a lower bound for $\mathrm{Vol}_d\left(\gamma \cap A\right)$. We will see that the latter also holds for $\gamma'$.

**Upper bound for** $\mathrm{Vol}_d(A)$. We refer to Figure 2 for the geometric constructions that follow. Let us first define $\Omega := \Delta(a, a') \cap A$ the intersection between the convex polytope $A$ and the hyperplane $\Delta(a, a')$. We also need to define $\Pi$ the set of all half-lines going from $a$ through $\Omega$, namely,

$$\Pi := \left\{ a + t(w - a) \mid \omega \in \Omega \text{ and } t > 0 \right\},$$

and the conic section $\Gamma := \mathcal{B}(a, \mathrm{diam}(A)) \cap \Pi$. We claim that $A \cap H_{a'} \subset \Gamma$. Indeed, let $\xi \in A \cap H_{a'}$. Since $\xi \in H_{a'}$, $[a, \xi]$ intersects $\Delta(a, a')$ in a unique point, say $\zeta$. By convexity, the segment $[a, \xi]$ is contained into $A$. In particular, $\zeta \in A$. Thus $\zeta \in \Delta(a, a') \cap A = \Omega$, and

$$\xi = a + \frac{\|\xi - a\|}{\|\zeta - a\|}(\zeta - a) \in A.$$

Moreover, since $\xi \in A$,

$$\|a - \xi\| \leq \sup_{s \in A} \|a - s\| = \mathrm{diam}(A),$$

and $\xi \in \mathcal{B}(a, \mathrm{diam}(A))$. A similar reasoning shows that $A \cap H_a \subset \Gamma'$, where $\Gamma'$ is the symmetric of $\Gamma$ with respect to $\Delta(a, a')$. Therefore,

$$\mathrm{Vol}_d(A) \leq 2\,\mathrm{Vol}_d(\Gamma).$$

Define the hyperspherical cap $\Sigma := \mathcal{S}(a, \rho) \cap \Pi$. Then we can express the volume of the conic section $\Gamma$ as

$$\mathrm{Vol}_d(\Gamma) = \frac{\mathrm{Vol}_{d-1}(\Sigma)}{\mathrm{Vol}_{d-1}(\mathcal{S}(a, \rho))}\,\mathrm{Vol}_d(\mathcal{B}(a, \mathrm{diam}(A))),$$

which leads to

$$\mathrm{Vol}_d(A) \leq \frac{2\,\mathrm{Vol}_{d-1}(\Sigma)}{\mathrm{Vol}_{d-1}(\mathcal{S}(a, \rho))}\,\mathrm{Vol}_d(\mathcal{B}(a, \mathrm{diam}(A))). \tag{2.2}$$

**Lower bound for** $\mathrm{Vol}_d(\gamma \cap A)$. By convexity, $\gamma \cap \Pi \subset \gamma \cap A$. Moreover,

$$\mathrm{Vol}_d(\gamma \cap \Pi) = \frac{\mathrm{Vol}_{d-1}(\Sigma)}{\mathrm{Vol}_{d-1}(\mathcal{S}(a, \rho))}\,\mathrm{Vol}_d(\mathcal{B}(a, \rho)).$$

Hence the following lower bound holds:

$$\mathrm{Vol}_d(\gamma \cap A) \geq \frac{\mathrm{Vol}_{d-1}(\Sigma)}{\mathrm{Vol}_{d-1}(\mathcal{S}(a, \rho))}\,\mathrm{Vol}_d(\mathcal{B}(a, \rho)). \tag{2.3}$$

**Conclusion.** Putting together Eq. (2.2) and (2.3), we obtain

$$\mathbb{P}(X_1 \in \gamma) \geq \frac{f_{\min}}{f_{\max}}\frac{\mathrm{Vol}_d(\mathcal{B}(a, \rho))}{\mathrm{Vol}_d(\mathcal{B}(a, \mathrm{diam}(A)))} = \frac{f_{\min}}{f_{\max}}\left(\frac{\rho}{\mathrm{diam}(A)}\right)^d.$$

Since $\rho \geq (c_r - 10)/(4Dc_r)$ and $\mathrm{diam}(A) \leq D$, we deduce that

$$\mathbb{P}(X_1 \in \gamma) \geq \frac{f_{\min}}{f_{\max}}\left(\frac{c_r - 10}{4c_r}\right)^d.$$

We conclude the proof by using the preliminary computations. □

## 2.2 Bad cuts

We now focus on the probability of a "bad split," that is, $\Delta(x, y)$ intersects both $\mathcal{B}(z, r)$ and $\mathcal{B}(z', r)$. The following result is an analogous of Lemma 11 in Dasgupta and Freund (2008).

**Lemma 2.2** (**Probability of bad cut is upper bounded**). *Suppose that assumption 1.1 holds. Let $A \subset \mathbb{R}^d$ be a full-dimensional convex polytope such that $\operatorname{diam}(A) \leq D < +\infty$. Let $c_r > 10$ be a constant. Pick any two balls $B := \mathcal{B}(z, r)$ and $B' := \mathcal{B}(z', r)$ such that*

- *(i) both $B$ and $B'$ intersect $A$;*

- *(ii) their radius is at most $D/c_r$;*

- *(iii) the distance between their centers satisfies $\|z - z'\| \geq D/2 - r$.*

*Then, if $X_1$ and $X_2$ are chosen independently from $A$ according to the distribution of $X$,*

$$\mathbb{P}\left(A \cap B \cap \Delta(X_1, X_2) \neq \emptyset \text{ and } A \cap B' \cap \Delta(X_1, X_2) \neq \emptyset\right) \leq \frac{f_{\max}}{f_{\min}} \frac{32 d c_r}{(c_r - 2)(c_r - 6)}.$$

*As a direct consequence, if $c_r > 15$,*

$$\mathbb{P}\left(A \cap B \cap \Delta(X_1, X_2) \neq \emptyset \text{ and } A \cap B' \cap \Delta(X_1, X_2) \neq \emptyset\right) \leq \frac{f_{\max}}{f_{\min}} \frac{64 d}{c_r}.$$

Note that, as in Lemma 2.1, the bound we obtain worsens as the dimension increases.

*Proof.* We first restrict ourselves to the case where the centers of $B$ and $B'$ both belong to $A$ with the same argument than in the proof of Lemma 2.1. Namely, define $a := \pi_A(z)$, $a' := \pi_A(z')$, $\beta := \mathcal{B}(a, r)$, $\beta' := \mathcal{B}(a', r)$. According to Lemma 2.3, $A \cap B \subset \beta$ and $A \cap B' \subset \beta'$. Thus

$$\mathbb{P}\left(A \cap B \cap \Delta(X_1, X_2) \neq \emptyset \text{ and } A \cap B' \cap \Delta(X_1, X_2) \neq \emptyset\right)$$
$$\leq \mathbb{P}\left(\beta \cap \Delta(X_1, X_2) \neq \emptyset \text{ and } \beta' \cap \Delta(X_1, X_2) \neq \emptyset\right).$$

For any $x \in \mathbb{R}^d$, define $B_x$ the set of points $y$ such that $\Delta(x, y)$ is a bad cut, that is,

$$B_x := \left\{y \in \mathbb{R}^d \mid \beta \cap \Delta(x, y) \neq \emptyset \text{ and } \beta' \cap \Delta(x, y) \neq \emptyset\right\}.$$

Then, since we sample $X_1$ according to the law of $X$ restricted to $A$ and since we assume Assumption 1.1 to be true,

$$\mathbb{P}\left(\beta \cap \Delta(X_1, X_2) \neq \emptyset \text{ and } \beta' \cap \Delta(X_1, X_2) \neq \emptyset\right) \leq \frac{f_{\max}}{f_{\min}} \frac{\mathbb{E}\left[\operatorname{Vol}_d\left(B_{X_1} \cap A\right)\right]}{\operatorname{Vol}_d(A)},$$

where the expectation is relative to the random variable $X_1$.

**Upper bound for** $\operatorname{Vol}_d(B_x \cap A)$**.** Let $x \in A$ and $y \in B_x$. By Lemma 2.4,

$$\begin{cases} (\|x - a\| - 2r)^+ & \leq \|y - a\| \leq \|x - a\| + 2r \\ (\|x - a'\| - 2r)^+ & \leq \|y - a'\| \leq \|x - a'\| + 2r. \end{cases}$$

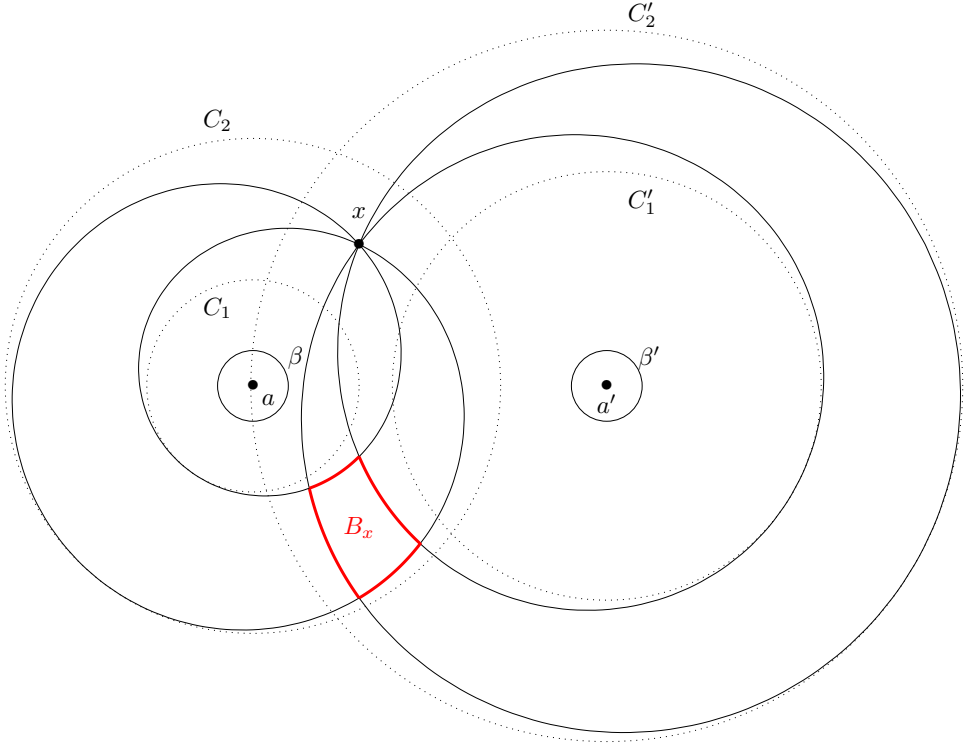

Figure 3: Sketch of $B_x$ in $\mathbb{R}^2$. For a fixed $x$, $B_x$ is the set of all $y$ such that $\Delta(x, y)$ cuts both $\beta$ and $\beta'$ (border marked in red). We show that $B_x$ is the intersection of two geometric loci (solid lines border). In particular, $B_x$ is included in the intersection of two annuli (border in dotted lines).

Equivalently, $B_x \subset \mathcal{A}(a, r_1, r_2) \cap B_x \subset \mathcal{A}(a', r_1', r_2')$, where we defined $r_1 := \|x - a\| - 2r$, $r_2 := \|x - a\| + 2r$, $r_1' := \|x - a'\| - 2r$ and $r_2' := \|x - a'\| + 2r$. Recall that $\mathcal{A}(a, r_1, r_2) = \mathcal{B}(a, r_2)$ whenever $r_1 \leq 0$. See Figure 3 for an illustration.

For any $\xi \in (a, a')$, denote by $D_\xi$ the hyperplane orthogonal to $(a, a')$ and passing through $\xi$. According to Lemma 2.6, the width of $\mathcal{A}(a, r_1, r_2) \cap \mathcal{A}(a', r_1', r_2')$ along the $(a, a')$ axis is upper bounded by $16D/(c_r - 2)$, hence there exists $\xi^-$ and $\xi^+ \in (a, a')$ such that $\|\xi^+ - \xi^-\| \leq 16D/(c_r - 2)$ and $B_x \cap A$ is contained between $D_{\xi^-}$ and $D_{\xi^+}$. For each $\xi \in (a, a')$, set $\Omega_\xi := D_\xi \cap A$. There exists $\xi^\star \in [\xi^-, \xi^+]$ such that $\mathrm{Vol}_{d-1}(\Omega_{\xi^\star})$ is maximal, and

$$
\begin{aligned}
\mathrm{Vol}_d(B_x \cap A) &= \int_{\xi \in [\xi^-, \xi^+]} \mathrm{Vol}_{d-1}(\Omega_\xi) \, \mathrm{d}\xi \\
&\leq \|\xi^+ - \xi^-\| \cdot \mathrm{Vol}_{d-1}(\Omega_{\xi^\star}) \\
\mathrm{Vol}_d(B_x \cap A) &\leq \frac{16}{c_r - 2} \cdot D \, \mathrm{Vol}_{d-1}(\Omega_{\xi^\star}) .
\end{aligned}
\tag{2.4}
$$

**Lower bound for** $\mathrm{Vol}_d(A)$. Suppose that $\xi^\star$ belongs to the segment $[a, a']$. By convexity, $A$ contains the (disjoint) union of the two hyperpyramids of apexes $a$ and $a'$ with $(d-1)$-dimensional basis $\Omega_{\xi^\star}$, which we denote by $Q$ and $Q'$. Therefore,

$$
\begin{aligned}
\mathrm{Vol}_d(A) &\geq \mathrm{Vol}_d(Q) + \mathrm{Vol}_d(Q') \\
&= \frac{\|a - \xi^\star\| \, \mathrm{Vol}_{d-1}(\Omega_{\xi^\star})}{d} + \frac{\|a' - \xi^\star\| \, \mathrm{Vol}_{d-1}(\Omega_{\xi^\star})}{d} \\
&= \frac{\delta \, \mathrm{Vol}_{d-1}(\Omega_{\xi^\star})}{d} \, .
\end{aligned}
$$

Since $\delta \geq (c_r - 6)D/(2c_r)$,

$$
\mathrm{Vol}_d(A) \geq \frac{c_r - 6}{2dc_r} \cdot D \, \mathrm{Vol}_{d-1}(\Omega_{\xi^\star}) \, . \tag{2.5}
$$

A similar reasoning holds whenever $\xi^\star$ does not belong to $[a, a']$.

**Conclusion.** Putting together Eq. (2.4) and (2.5), we obtain

$$
\mathbb{P}\left(\beta \cap \Delta(X_1, X_2) \neq \emptyset \text{ and } \beta' \cap \Delta(X_1, X_2) \neq \emptyset\right) \leq \frac{f_{\max}}{f_{\min}} \frac{32dc_r}{(c_r - 2)(c_r - 6)} \, ,
$$

which concludes the proof. □

Note that in the plane defined by $a$, $a'$ and $x$, we can actually describe precisely the shape of the curves defining the border of $B_x$—see Figure 3. These curves correspond to the images of $x$ by all the symmetries with respect to a line tangent to $\beta$ or $\beta'$. Individually, they are called the *orthotomics of a circle*, or second caustic (Lawrence, 2013, p. 60).

## 2.3   Technical results

This first lemma is used in the proofs of Lemma 2.1 and 2.2 to deal with cases where the center of $B$ or $B'$ does not belong to $A$. See Figure 4 for an illustration of such a situation.

**Lemma 2.3** (Construction of $\beta$). *Let $A \subset \mathbb{R}^d$ be a convex compact set and $\mathcal{B}(z, r)$ be a ball that intersects $A$. Define $\beta := \mathcal{B}(\pi_A(z), r)$. Then $A \cap B \subset \beta$.*

*Proof.* Set $a := \pi_A(z)$. Let $x$ be an element of $A \cap B$. Then,

$$
\begin{aligned}
\|x - a\|^2 &= \langle x - a, x - a \rangle \\
&= \langle x - z + z - a, x - z + z - a \rangle \\
&= \|x - z\|^2 + 2\langle x - z, z - a \rangle + \|z - a\|^2 \\
\|x - a\|^2 &= \|x - z\|^2 + 2\langle x - a, z - a \rangle - \|z - a\|^2 \, .
\end{aligned}
$$

Since $\pi_A$ is the orthogonal projection, given that $x \in A$, we have $\langle x - a, z - a \rangle \leq 0$. Moreover, $\|z - a\| \geq 0$, thus $\|x - a\|^2 \leq \|x - z\|^2$. But $x$ also belongs to $B$, hence $\|x - z\| \leq r$. As a consequence, $\|x - a\| \leq r$, that is, $x \in \beta$. □

The next lemma shows that, for a given $x$, the set of every possible $y$ such that $\Delta(x, y)$ intersects $\mathcal{B}(a, r)$ is contained into an annulus centered in $a$. We refer to Figure 5 for an illustration.

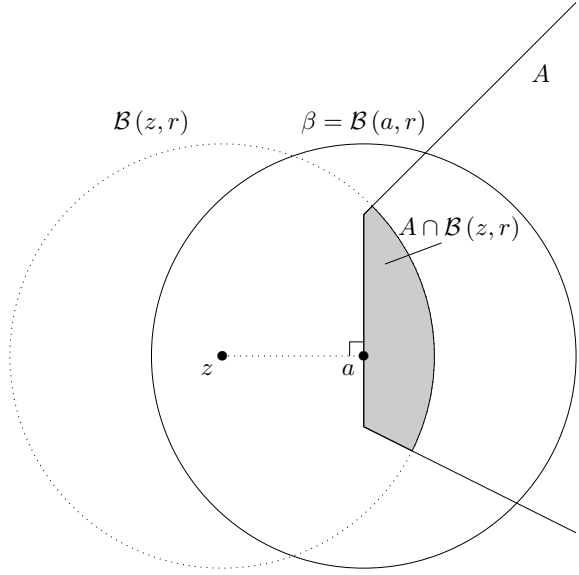

Figure 4: Construction of $\beta$. The point $a$ is the image of $z$ by the orthogonal projection on $A$. The ball $\beta$ has the same radius than $\mathcal{B}(z, r)$ and contains $A \cap \mathcal{B}(z, r)$, which is marked in gray.

**Lemma 2.4** (**Localization of $B_x$**). *Let $a, x \in \mathbb{R}^d$ and $r > 0$. Then, for any $y \in \mathbb{R}^d$ such that $\Delta(x, y) \cap \mathcal{B}(a, r)$ is non-empty,*

$$(\|x - a\| - 2r)^+ \le \|y - a\| \le \|x - a\| + 2r\,.$$

*Proof.* Let $y \in \mathbb{R}^d$ such that $\Delta(x, y) \cap \mathcal{B}(a, r)$ is non-empty. In particular, there exists $b \in \mathbb{R}^d$ such that $\|y - b\| = \|x - b\|$ and $\|a - b\| \le r$. By the triangle inequality,

$$\big|\|y - a\| - \|y - b\|\big| \le \|a - b\| \le r\,.$$

Hence

$$\begin{cases} \|y - a\| \le r + \|y - b\| = r + \|x - b\| \\ \|y - a\| \ge -r + \|y - b\| = -r + \|x - b\|\,. \end{cases}$$

Since $\big|\|x - b\| - \|a - b\|\big| \le \|x - a\|$ (again by the triangle inequality), we have

$$\begin{cases} \|y - a\| \le \|x - a\| + 2r \\ \|y - a\| \ge \|x - a\| - 2r\,. \end{cases}$$

$\square$

We now present a result stating that, for any two points $a, a' \in \mathbb{R}^d$, there exists a simple set of possible $x$ and $y$ such that balls with center $a$ and $a'$ are well-separated by $\Delta(x, y)$. It is the key element in the proof of Lemma 2.1.

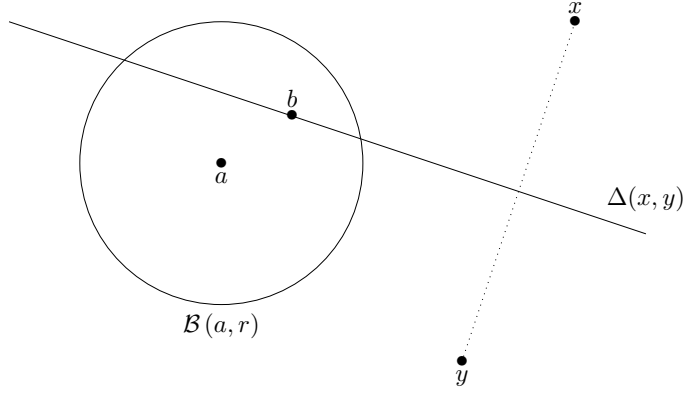

Figure 5: $B_x$ is included in the intersection of two annuli. As in the proof of Lemma 2.4, $a$ and $x$ are fixed, and $y$ is such that $\Delta(x, y)$ intersects $\mathcal{B}(a, r)$. Then $y$ belongs to an annulus of radii $(\|x - a\| - 2r)^+$ and $\|x - a\| + 2r$.

**Lemma 2.5 (Sufficient condition for a good cut).** *Let* $a, a' \in \mathbb{R}^d$. *Let* $0 < r < \|a - a'\| / 2$ *and set* $\rho := \|a - a'\| / 2 - r$. *Then, for any* $x \in \mathcal{B}(a, \rho)$ *and* $y \in \mathcal{B}(a', \rho)$, *we have* $\mathcal{B}(a, r) \subset H_x$ *and* $\mathcal{B}(a', r) \subset H_y$.

*Remark* 2.1. Note that Lemma 2.5 holds in any metric space $(\mathcal{X}, \delta)$ since the proof only uses the triangle inequality.

*Proof.* We refer to Figure 6 for this proof. We have to prove that for any $s \in \mathcal{B}(a, r)$, $\delta(s, x) \leq \delta(s, y)$ (the case $t \in \mathcal{B}(a', r)$ is identical up to notations). We first write

$$\delta(s, x) \leq \delta(s, a) + \delta(a, x) \leq r + \rho = \delta(a, a') / 2,$$

where we used (i) the triangle inequality, (ii) $s \in \mathcal{B}(a, r)$ and $x \in \mathcal{B}(a, \rho)$, (iii) the definition of $\rho$. Then,

$$\delta(a, a') \leq \delta(a, y) + \delta(a', y) \leq \delta(a, y) + \rho,$$

where we used (i) triangle inequality, (ii) $y \in \mathcal{B}(a', \rho)$. Thus $\delta(a, y) \geq \delta(a, a') - \rho$. Moreover,

$$\delta(a, y) \leq \delta(a, s) + \delta(s, y) \leq r + \delta(s, y),$$

where we used (i) triangle inequality, (ii) $s \in \mathcal{B}(a, r)$. Combining the two, we get

$$\delta(s, y) \geq \delta(a, a') - (r + \rho) = \delta(a, a') / 2.$$

Therefore $\delta(s, y) \geq \delta(s, x)$ and we can conclude. $\qquad\square$

Finally, we state and prove a technical lemma used in the proof of Lemma 2.2 to control the size of the intersection of two annuli.

**Lemma 2.6 ($B_x$ has small width).** *Assume the set of hypotheses of Lemma 2.2 and define* $r_1$, $r_2$, $r_1'$ *and* $r_2'$ *as in the proof of Lemma 2.2. Then there exist two hyperplanes* $L_x$

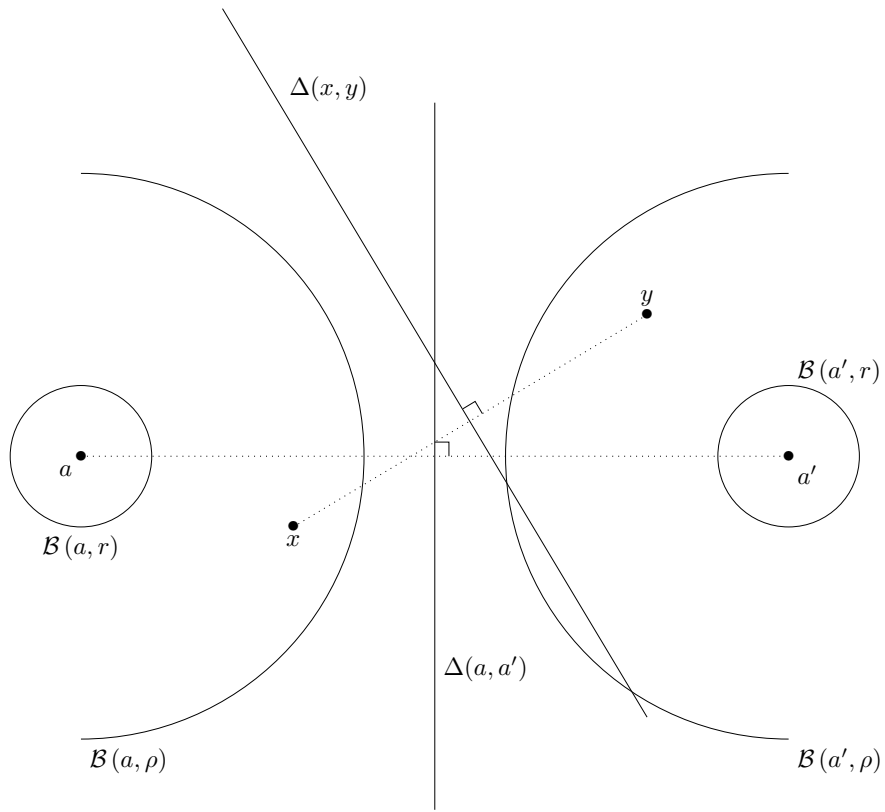

Figure 6: Guaranteed good cut. Set $a, a' \in \mathbb{R}^d$ and $\rho = \|a - a'\| /2 - r$. Then, for any $x \in \mathcal{B}(a, \rho)$ and $y \in \mathcal{B}(a', \rho)$, the hyperplane $\Delta(x, y)$ separates cleanly $\mathcal{B}(a, r)$ from $\mathcal{B}(a', r)$.

and $L'_x$, orthogonal to $(a, a')$, such that the intersection of $\mathcal{A}(a, r_1, r_2)$ and $\mathcal{A}(a', r'_1, r'_2)$ is included between $L_x$ and $L'_x$. Additionally,

$$\delta\left(L_x, L'_x\right) \leq \frac{16D}{c_r - 2}. \tag{2.6}$$

Even though the statement of Lemma 2.6 may seem intuitive at first sight (since the intersection is contained in two annuli of width $O\left(D/c_r\right)$, one would expect its width to be of the same order), we do not know of a simpler proof. We believe that it is necessary to describe precisely the intersection of the two annuli depending on the radii in order to make sure that the situation where the two annuli are overlapping is excluded. Indeed, in this case the width of the intersection is *not* bounded by a quantity depending on $D/c_r$ but rather on $D$, since it has the same order than the diameter of the annuli.

*Proof.* By rotational symmetry around $(a, a')$, it suffices to prove the result in a 2-plane containing $a$ and $a'$. Hence from now on we work in the plane $P$ defined by the triple $(x, a, a')$. We first describe the shape of the intersection between the two annuli depending on the position of $x$ relatively to $a$ and $a'$. Then, in each case, we bound the width of the intersection in the direction of the $(a, a')$ axis.

**Shape of $\mathcal{A}(a, r_1, r_2) \cap \mathcal{A}(a', r'_1, r'_2)$.** Let us equip $P$ with an orthogonal frame such that $a = (0, 0)$, $a' = (\delta, 0)$ and $x = (x_1, x_2)$. The width of the intersection is invariant by line symmetry with respect to $\Delta(a, a')$ and $(a, a')$, thus we can restrict our analysis to the quadrant defined by $x_1 \leq \delta/2$ and $x_2 > 0$. In particular, $\|x - a\| \leq \|x - a'\|$. Define $C_i := \mathcal{C}(a, r_i)$ and $C'_i := \mathcal{C}(a', r'_i)$ for $i \in \{1, 2\}$. The shape of $\mathcal{A}(a, r_1, r_2) \cap \mathcal{A}(a', r'_1, r'_2)$ depends on the mutual intersections between $C_1, C_2, C'_1$ and $C'_2$. Recall that $\mathcal{C}(a, \rho) \cap \mathcal{C}(a', \rho') \neq \emptyset$ if, and only if,

$$|\rho - \rho'| \leq \delta \leq \rho + \rho'.$$

We now proceed to describe these intersection depending on the position of $x$ relatively to $a$ and $a'$.

- Since $r > 0$, $r_1 < r_2$ and $r'_1 < r'_2$ and thus $C_1 \cap C_2 = C'_1 \cap C'_2 = \emptyset$;

- By the triangle inequality, $|r_2 - r'_2| = |\|x - a\| - \|x - a'\|| \leq \delta$ and $r_2 + r'_2 = \|x - a\| + \|x - a'\| + 4r \geq \delta$, hence $C_2 \cap C'_2$ is always non-empty;

- By the triangle inequality, $|r_1 - r'_1| = |\|x - a\| - \|x - a'\|| \leq \delta$. Hence $C_1 \cap C'_1$ is non-empty if, and only if, $r_1 + r'_1 \geq \delta$, that is, $\|x - a\| + \|x - a'\| \geq \delta - 4r$. The border is an ellipse with focal points $a, a'$ and semi-major axis $(\delta + 4r)/2$.

- By the triangle inequality, $r_1 + r'_2 = \|x - a\| + \|x - a'\| \leq \delta$. Since $\|x - a\| \leq \|x - a'\|$, $|r_1 - r'_2| = 4r - \|x - a\| + \|x - a'\|$. Thus $C_1 \cap C'_2$ is non-empty if, and only if, $\|x - a'\| - \|x - a\| \leq \delta - 4r$. The border is a hyperbola with focal points $a, a'$ and semi-major axis $(\delta - 4r)/2$.

- By the triangle inequality, $r'_1 + r_2 = \|x - a'\| + \|x - a\| \geq \delta$. Moreover, $|r'_1 - r_2| = |\|x - a'\| - \|x - a\| - 4r|$. Again, the triangle inequality yields $\|x - a'\| - \|x - a\| \leq \delta \leq \delta + 4r$. On the other side, $\|x - a\| - \|x - a'\| \leq 0 \leq \delta - 4r$ because $r < \delta/4$. Hence $C_2 \cap C'_1$ is always non-empty.

The different cases are summarized in Figure 7, and we provide a visual depiction of the intersection for each case in Figure 8. Note that in Case III, $\|x - a\| \leq 2r$ is a possibility, implying $r_1 < 0$. In this event, we see in Figure 8 that the extremal points are the same.

14

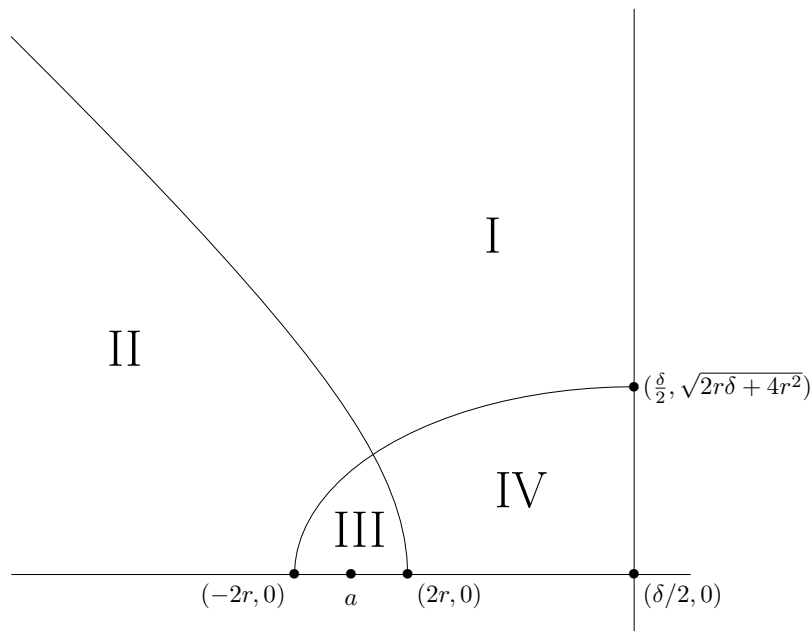

Figure 7: Shape of $\mathcal{A}(a, r_1, r_2) \cap \mathcal{A}(a', r_1', r_2')$ (I). Depending on the relative position of $x$ with respect to $a$ and $a'$, the shape of $\mathcal{A}(a, r_1, r_2) \cap \mathcal{A}(a', r_1', r_2')$ changes. *Case I:* $C_1 \cap C_1'$ and $C_1 \cap C_2'$ are both non-empty. *Case II:* $C_1 \cap C_1'$ is non-empty, whereas $C_1 \cap C_2'$ is. *Case III:* $C_1 \cap C_1'$ and $C_1 \cap C_2'$ are both empty. *Case IV:* $C_1 \cap C_2'$ is non-empty whereas $C_1 \cap C_1'$ is. The shape of $B_x$ as well as $\mathcal{A}(a, r_1, r_2) \cap \mathcal{A}(a', r_1', r_2')$ in this last case is depicted in Figure 3.

15

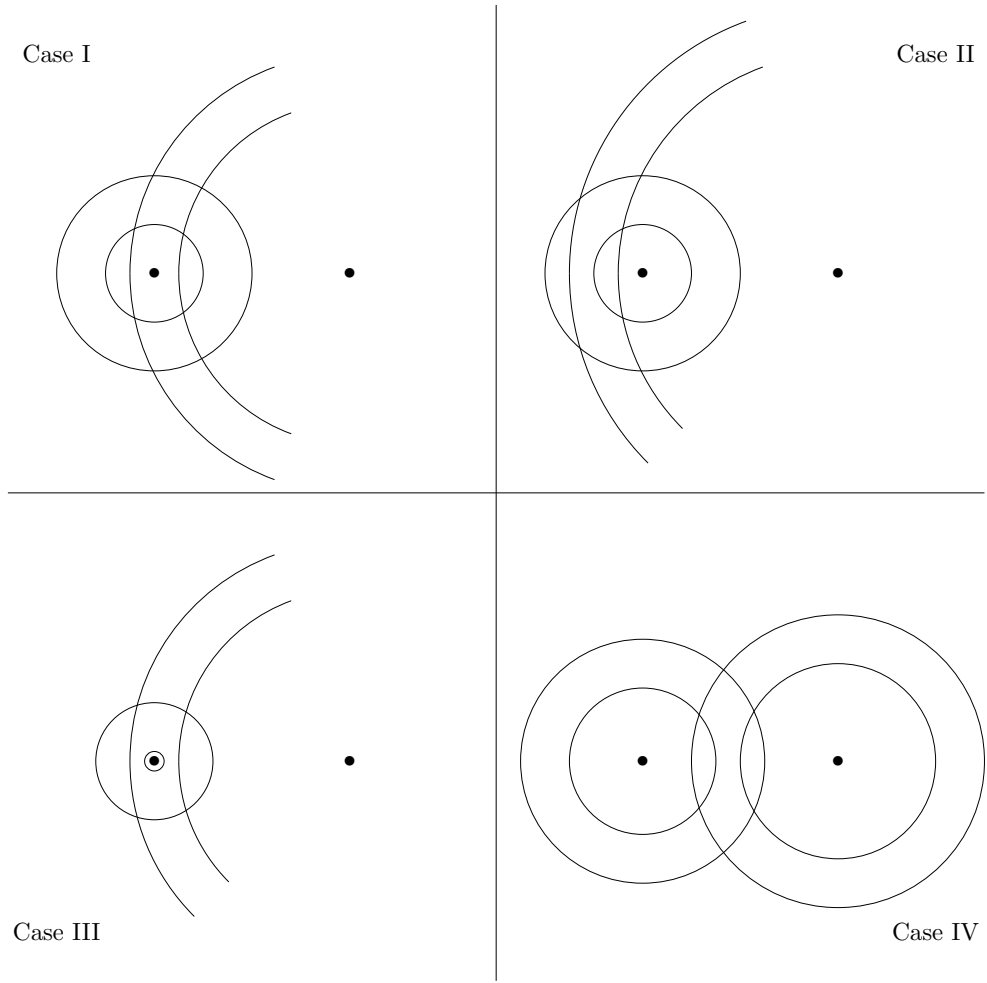

Figure 8: Shape of $\mathcal{A}(a, r_1, r_2) \cap \mathcal{A}(a', r_1', r_2')$ (II). For each case described in Figure 7, we sketch $\mathcal{A}(a, r_1, r_2) \cap \mathcal{A}(a', r_1', r_2')$. Note that the points realizing the minimum and maximum abscissa in each case are different, leading to different bounds on the width of $\mathcal{A}(a, r_1, r_2) \cap \mathcal{A}(a', r_1', r_2')$.

**Width of** $\mathcal{A}(a, r_1, r_2) \cap \mathcal{A}(a', r_1', r_2')$**.** For each case, we show that Eq. (2.6) holds. Recall that we assumed $r/\delta \leq 2/(c_r - 2)$ and $\operatorname{diam}(A) \leq D$. We will use the fact that

$$\frac{\|x - a\|^2 - \|x - a'\|^2}{2\delta} = \frac{x_1^2 + x_2^2 - x_1^2 + 2\delta x_1 + \delta^2 + x_2^2}{2\delta} = x_1 - \frac{\delta}{2}.$$

- Case I: The left-most points of $\mathcal{A}(a, r_1, r_2) \cap \mathcal{A}(a', r_1', r_2')$ belong to $C_1 \cap C_2'$. We solve

$$\begin{cases} \xi_1^2 + \xi_2^2 & = r_1^2 = (\|x - a\| - 2r)^2 \\ (\xi_1 - \delta)^2 + \xi_2^2 & = r_2'^2 = (\|x - a'\| + 2r)^2 \,. \end{cases}$$

 and find

$$\xi_1 = x_1 - \frac{2r}{\delta} \left( \|x - a\| + \|x - a'\| \right).$$

 The right-most points of $\mathcal{A}(a, r_1, r_2) \cap \mathcal{A}(a', r_1', r_2')$ belong to $C_2 \cap C_1'$. We solve

$$\begin{cases} \zeta_1^2 + \zeta_2^2 & = r_2^2 = (\|x - a\| + 2r)^2 \\ (\zeta_1 - \delta)^2 + \zeta_2^2 & = r_1'^2 = (\|x - a'\| - 2r)^2 \,. \end{cases}$$

 and find

$$\zeta_1 = x_1 + \frac{2r}{\delta} \left( \|x - a\| + \|x - a'\| \right).$$

 Thus the width of $\mathcal{A}(a, r_1, r_2) \cap \mathcal{A}(a', r_1', r_2')$ along $(a, a')$ is given by

$$|\zeta_1 - \xi_1| = \frac{4r}{\delta} \left( \|x - a\| + \|x - a'\| \right) \leq \frac{16D}{c_r - 2}.$$

- Case II: The left-most point of $\mathcal{A}(a, r_1, r_2) \cap \mathcal{A}(a', r_1', r_2')$ belongs to $C_2' \cap (a, a')$, and we have

$$\xi_1 = \delta - r_2' = \delta - \|x - a'\| - 2r.$$

 The right-most points belongs to $C_2 \cap C_1'$, and we have, as in Case I,

$$\zeta_1 = x_1 + \frac{2r (\|x - a\| + \|x - a'\|)}{\delta}.$$

 Thus the width of $\mathcal{A}(a, r_1, r_2) \cap \mathcal{A}(a', r_1', r_2')$ along $(a, a')$ is given by

$$|\zeta_1 - \xi_1| = \frac{2r}{\delta} \left( \|x - a\| + \|x - a'\| \right) + \|x - a'\| + x_1 - \delta - 2r.$$

 The equation of the asymptotes of the hyperbola $\|x - a'\| - \|x - a\| = \delta - 4r$ are given by

$$x_2 = \pm \frac{2\sqrt{2r\delta - 4r^2}}{\delta - 4r}(x_1 - \delta/2),$$

 and considering the lines parallel to these asymptotes passing through $(\delta, 0)$ we deduce that, in case II,

$$\frac{x_2^2}{(x_1 - \delta)^2} \leq \frac{4(2r\delta - 4r^2)}{(\delta - 4r)^2} \leq \frac{8r\delta}{(\delta - 4r)^2} \leq 8 \frac{r}{\delta} \frac{1}{\left(1 - 4\frac{r}{\delta}\right)^2} \leq \frac{16 (c_r - 10)^2}{(c_r - 2)^3} \,.$$

Thus

$$\|x - a'\| = \sqrt{(x_1 - \delta)^2 + x_2^2} \leq |x_1 - \delta| \sqrt{1 + \frac{16\,(c_r - 10)^2}{(c_r - 2)^3}},$$

and we have

$$\|x - a'\| + x_1 - \delta \leq |x_1 - \delta| \left( \sqrt{1 + \frac{16\,(c_r - 10)^2}{(c_r - 2)^3}} - 1 \right) \leq \frac{8D\,(c_r - 10)^2}{(c_r - 2)^3},$$

where we used $\sqrt{1 + x^2} - 1 \leq x/2$ in the last inequality. Finally,

$$|\zeta_1 - \xi_1| \leq \frac{8D(c_r - 10)^2}{(c_r - 2)^3} + \frac{8D}{c_r - 2}.$$

- Case III: The left-most point of $\mathcal{A}\,(a, r_1, r_2) \cap \mathcal{A}\,(a', r_1', r_2')$ belongs to $C_2 \cap C_2'$. We solve

$$\begin{cases} \xi_1^2 + \xi_2^2 & = r_2^2 = (\|x - a\| + 2r)^2 \\ (\xi_1 - \delta)^2 + \xi_2^2 & = r_2'^2 = (\|x - a'\| + 2r)^2, \end{cases}$$

which yields

$$\xi_1 = x_1 + \frac{2r}{\delta}\,(\|x - a\| - \|x - a'\|).$$

The right-most points belongs to $C_2 \cap C_1'$, and we have, as in Case I,

$$\zeta_1 = x_1 + \frac{2r\,(\|x - a\| + \|x - a'\|)}{\delta}.$$

Thus the width of $\mathcal{A}\,(a, r_1, r_2) \cap \mathcal{A}\,(a', r_1', r_2')$ along $(a, a')$ is given by

$$|\zeta_1 - \xi_1| = \frac{4r}{\delta}\,\|x - a'\| \leq \frac{8D}{c_r - 2}.$$

- Case IV: as in Case I, the width of $\mathcal{A}\,(a, r_1, r_2) \cap \mathcal{A}\,(a', r_1', r_2')$ is given by

$$|\xi_1 - \zeta_1| = \frac{4r\,(\|x - a\| + \|x - a'\|)}{\delta}.$$

Since in this case $\|x - a\| + \|x - a'\| \leq \delta - 4r$, we have

$$|\xi_1 - \zeta_1| \leq 4r \leq \frac{4D}{c_r}.$$

Overall, since $c_r > 10$, we have shown that the width of $\mathcal{A}\,(a, r_1, r_2) \cap \mathcal{A}\,(a', r_1', r_2')$ along $(a, a')$ is upper bounded by $16D/(c_r - 2)$. $\qquad \square$

# 3 Additional experiments

## 3.1 Comparison of embedding methods

Here we report the results of the comparison between embedding methods. The results are provided as a supplement to Section 4.3. We use a subsample of $n = 500$ points

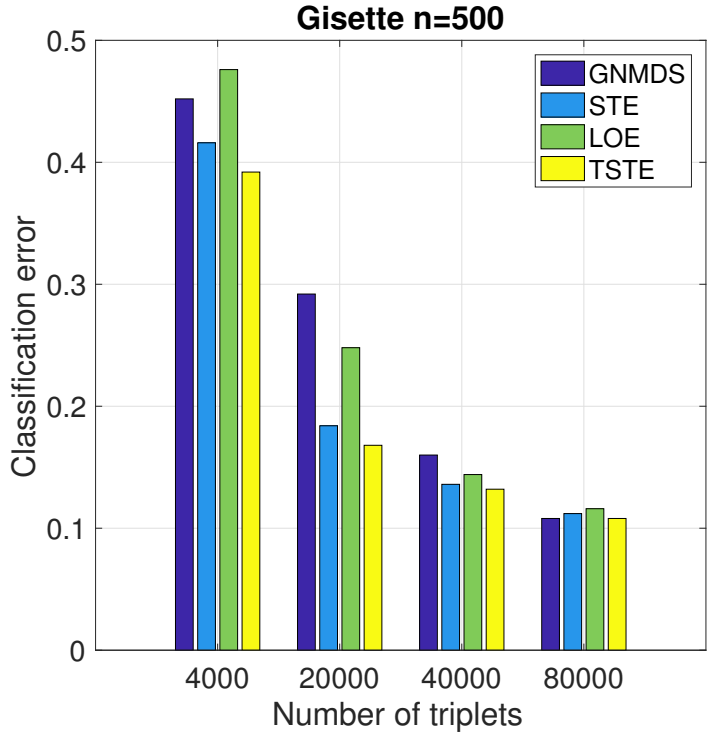

Figure 9: Classification error of the various embedding methods and the KNN algorithm.

from the Gisette dataset (half as training set and half as test set). The dimension $d \in \{10, 20, 30, 40, 50\}$ and $k \in \{2, 6, 10, 14, 18\}$ of the KNN are adjusted with 2-fold cross-validation on the training set. Figure 9 shows the classification error of the four embedding methods: GNMDS (Agarwal et al., 2007), LOE (Terada and von Luxburg, 2014) and STE/TSTE (van der Maaten and Weinberger, 2012).

The TSTE consistently outperforms other methods. Therefore, we use it as the main competitor against our proposed random forest.

## 3.2 CompRF and subsampling

In this section we investigate the role of subsampling in the performance of the CompRF. To construct each tree of the CompRF, we randomly pick a subsample of $r|S|$ points among the set of training points ($S$) without replacement and make the tree only based on the subsample. We use the following range for the subsampling ratio: $r \in \{0.1, 0.2, 0.4, 1\}$. The left panel of Figure 10 shows the average classification error of the CompRF for various values of $r$. The right plot in this figure shows the normalized average MSE of the CompRF for regression datasets. Note that the range of MSE depends on the dataset. To make a unified figure, for each dataset, we divided all average values of the MSE by the maximum value of the MSE on that particular dataset.

Our results hardly show any significant positive effect of subsampling. On the contrary, in

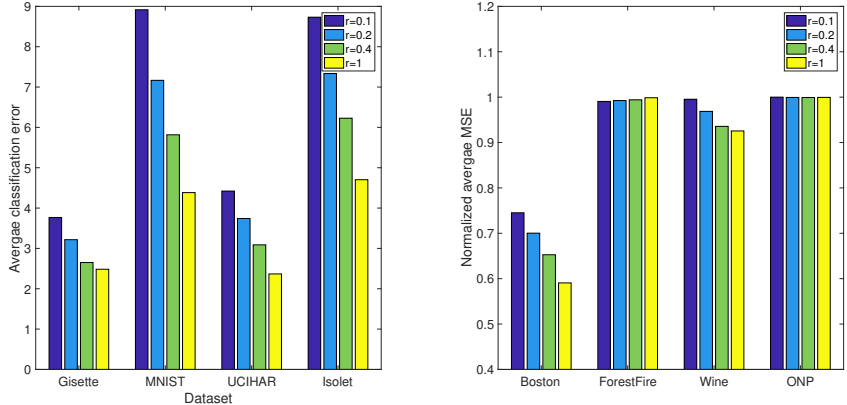

Figure 10: (Left) Average classification error of the CompRF algorithm on four classification datasets and various subsampling ratios ($r$). (Right) Normalized average MSE of the CompRF algorithm on four regression datasets and various subsampling ratios ($r$). The X-axis denotes the datasets. Note that for each dataset we divided all MSE values by the maximum value of the dataset. In this way bars can be plotted together.

classification tasks we see a significant decrease in error when the whole dataset is used. Only in case of ForestFire dataset do we see some slight improvement.

## 3.3 Running time of CompRF vs. Embedding procedures

Here we report the running time of CompRF in comparison with TSTE embedding combined with KNN. Note that if we apply CART forest after embedding, it can be even more time consuming. In addition, the running time of embedding does not change significantly if we apply the same triplets as the CompRF or a random subsample of triplets, therefore we report the running time based on the same triplets as the CompRF.

We use the subsample of Gisette dataset with $n = 1000$ point, similar to the Section 4.3. We perform the embedding with $d = 10$ and $d = 50$ dimensions and fixed $k = 5$. Table 1 shows the running time of the experiments. Since the running time of embedding can change significantly based on the initial conditions, we run embedding algorithms five times and we report the average running time. The algorithms are implemented on a single core CPU and the running times are reported in seconds.

Table 1: Comparison of computation time between CompRF and TSTE+KNN. The reported values are in seconds.

| Number of trees (M) | 1 | 5 | 10 | 20 |
|---|---|---|---|---|
| CompRF | 1 | 4 | 8 | 16 |
| TSTE+KNN (d=10) | 148 | 236 | 350 | 595 |
| TSTE+KNN (d=50) | 185 | 654 | 1214 | 2398 |

The required running time for the embedding algorithm is orders of magnitude longer than

the CompRF. Moreover, the embedding algorithms need a cross-validation step to adjust
the number of dimensions and other parameters of the classifier.



\end{document}